\pdfoutput=1

\documentclass[11pt]{article}

\usepackage[final]{acl}
\usepackage{booktabs}
\usepackage{multirow}

\usepackage{xurl}
\usepackage{times}
\usepackage{latexsym}
\usepackage{adjustbox}
\usepackage{amsmath} 
\usepackage[capitalise]{cleveref}

\usepackage[T1]{fontenc}

\usepackage[utf8]{inputenc}

\usepackage{microtype}
\usepackage{algorithm, algpseudocode}
\usepackage{inconsolata}
\usepackage{subcaption}
\usepackage{graphicx}
\usepackage{url} 
\title{BrightCookies at SemEval-2025 Task 9: Exploring Data Augmentation for Food Hazard Classification}

\author{
 \textbf{Foteini Papadopoulou\textsuperscript{1,2}},
 \textbf{Osman Mutlu\textsuperscript{1}},
 \textbf{Neris Özen\textsuperscript{1}},\\
 \textbf{Bas H.M. van der Velden\textsuperscript{1}},
 \textbf{Iris Hendrickx\textsuperscript{2}},
 \textbf{Ali Hürriyetoğlu\textsuperscript{1}}
\\
\\
 \textsuperscript{1}Wageningen Food Safety Research, The Netherlands \\
 \textsuperscript{2}Centre for Language Studies, Radboud University, The Netherlands
\\
 \small{
   \textbf{Correspondence:} \href{mailto:ali.hurriyetoglu@wur.nl}{ali.hurriyetoglu@wur.nl}
 }
}
\usepackage{array}
\newcolumntype{P}[1]{>{\centering\arraybackslash}p{#1}}
\begin{document}
\maketitle
\begin{abstract}
This paper presents our system developed for the SemEval-2025 Task 9: The Food Hazard Detection Challenge. The shared task's objective is to evaluate explainable classification systems for classifying hazards and products in two levels of granularity from food recall incident reports. In this work, we propose text augmentation techniques as a way to improve poor performance on minority classes and compare their effect for each category on various transformer and machine learning models. We explore three word-level data augmentation techniques, namely synonym replacement, random word swapping, and contextual word insertion. The results show that transformer models tend to have a better overall performance. None of the three augmentation techniques consistently improved overall performance for classifying hazards and products. We observed a statistically significant improvement (P < 0.05) in the fine-grained categories when using the BERT model to compare the baseline with each augmented model. Compared to the baseline, the contextual words insertion augmentation improved the accuracy of predictions for the minority hazard classes by 6\%.
This suggests that targeted augmentation of minority classes can improve the performance of transformer models.
\end{abstract}

\section{Introduction}

Foodborne diseases affect millions of people every year. The World Health Organization highlights that food contamination leads to more than 200 diseases, resulting in severe health complications and affecting the socioeconomic stability of communities and nations \cite{WHO2024FoodSafety}. There is a vast amount of publicly available information on food safety-related websites. Given the importance of early detection of food hazards, there is a need to timely and accurately analyze all this publicly available information to detect food hazards.
\begin{figure}[ht]
    \centering
    \includegraphics[width=0.78\linewidth]{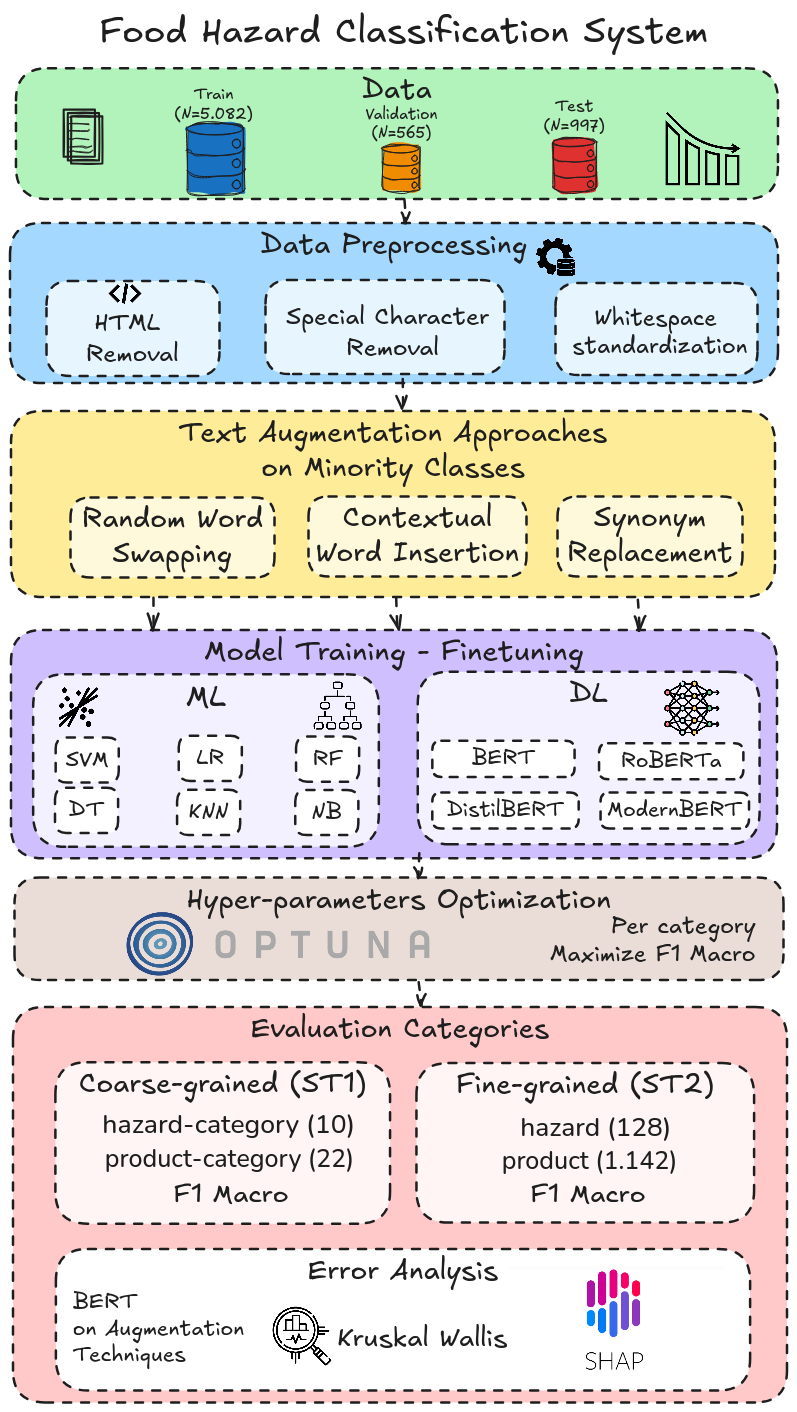}
    \caption{An overview of our developed system's architecture.}
    \label{fig:overview_arch}
    \vspace{-0.5cm}
\end{figure}

The SemEval-2025 Task 9: Food Hazard Detection Challenge \cite{semeval2025-task9} was proposed to facilitate automated classification of food hazards in food safety-related documents.  It stimulates research that combines food safety and natural language processing (NLP) for explainable multi-class classification of food recall incident reports. SemEval-2025 Task 9 includes two sub-tasks: classifying coarse food hazard and product categories (ST1) (\texttt{hazard-category, product-category}), and fine-grained hazard and product categories (ST2) (\texttt{hazard, product}).

A significant challenge with the SemEval-2025 Task 9 dataset is its substantial class imbalance. There is a long-tailed distribution across classes, especially in the fine-grained categories. This imbalance can give poor performance of classifiers, especially for deep learning (DL) models \cite{henning-etal-2023-survey}. Text augmentation techniques have been shown to mitigate the effects of imbalanced data to an extent \cite{10866632}.
 Text augmentation can range from simple string manipulations, such as those used in Easy Data Augmentation (EDA) \cite{wei-zou-2019-eda}, to more advanced methods involving transformer-based text generation \cite{henning-etal-2023-survey}. This helps to boost the representation of minority classes, which can result in a more balanced dataset and robust models.

We investigated three basic text augmentation techniques (synonym replacement, contextual word insertion, and random word swapping) to boost the representation of under-represented classes in multi-class classifications of food recall incident reports. Our main research question is: 

\textbf{Can text augmentation techniques on under-represented classes enhance a food hazard multi-class classifier's performance?}

We evaluated the performance of various machine learning (ML) algorithms and encoder-only transformer models, both in their baseline form and after applying each augmentation technique. To participate in the task, only one submission was allowed. 
We submitted our predictions in the official test set after we evaluated our models in the development set, selecting the best-performing ones for each category. In ST1, our system ranked 15\textsuperscript{th} out of 27 participants, with an $F_{1}$-macro score difference of 0.0613 from the first, and in ST2, it ranked 11\textsuperscript{th} out of 26, with a 0.0944 score gap from the top (see \autoref{results} for the exact scores). 
Our work provides valuable insights into the efficacy of text augmentation in this field.
\footnote{ Our code is available at \url{https://github.com/WFSRDataScience/SemEval2025Task9} }





\section{Related Work}
\subsection{Research on food hazard classification}
Little work has been conducted using text data for fine-grained food hazard classification \cite{randl-etal-2024-cicle}, as most existing literature focused on binary classification of food hazards. A recent study by \citet{randl-etal-2024-cicle} introduced the dataset that we used in SemEval-2025 Task 9 and they benchmarked multiple ML and DL algorithms. 
\citet{randl-etal-2024-cicle} proposed a large language model (LLM)-in-the-loop framework named Conformal In-Context Learning (CICLe), that leveraged Conformal Prediction to optimize context length for predictions of a base classifier. By using fewer, more targeted examples, performance increased and energy consumption reduced compared to regular prompting.


\subsection{Text augmentation for minority classes}
Data augmentation creates synthetic data from an existing dataset by inserting small changes into copies of the data \cite{Shorten2021}. Data augmentation mitigates the class imbalance issues for DL \cite{henning-etal-2023-survey}. According to \citet{Shorten2021}, Data augmentation approaches in NLP can be divided in two types: symbolic and neural. Symbolic techniques, such as rule-based EDA \cite{wei-zou-2019-eda}, employ simple word-level operations like synonym replacement and random insertion. Symbolic techniques are effective in small datasets. Neural techniques rely on auxiliary neural networks such as back-translation or generative augmentation. A recent study showed that LLMs for data augmentation, such as to generate new samples, increase accuracy and address class imbalance in skewed datasets \citep{Gopali}. In our study, we explore both symbolic and simple neural augmentation strategies, such as contextual words insertion using BERT, to improve classification performance.

Additionally, in the SemEval shared task of 2023, \citet{al-azzawi-etal-2023-nlp} explored the effects of data augmentation, particularly back translation, on minority classes. They compared it with augmenting the entire dataset using transformer-based models. They observed that targeting the underrepresented classes for augmentation proved more effective than broad dataset augmentation. Following their approach, we also focus our augmentation strategies on the minority classes rather than the entire dataset.

\section{Data}
The Food Recall Incidents dataset used in SemEval-2025 Task 9 contains 6,644 food-recall announcements in the English language \cite{randl_2024_10891602}. This dataset is split into 5,082 announcements in the train set, 565 in the development set, and 997 in the test set. The data is collected from 24 different websites (\autoref{tab:domain_samples}). 
The samples consist of a \texttt{title} and \texttt{text} describing announcements from a recalled food product and includes other metadata.

Experts manually labeled each sample into four coarse classes of hazards (\texttt{hazard-category}) and products (\texttt{product-category}) and fine-grained classes (\texttt{hazard} and \texttt{product}). The classes and the number of classes per category are listed in \autoref{tab:labels}, and examples are presented in \autoref{tab:examples}. 

The distribution of the four categories' classes is highly imbalanced, showing a long-tail effect (\autoref{fig:dist_cat}, \autoref{fig:dist_}). In coarse categories, 75\% of classes have 513 samples in \texttt{hazard-category} and 263 in \texttt{product-category}, while the largest classes contain 1,854 and 1,434 samples, respectively. This imbalance is even more severe in the fine-grained \texttt{hazard} and \texttt{product} classes, with 75\% of classes having at most four samples per \texttt{product} and 24 samples per \texttt{hazard}, while the largest class has 185 samples per \texttt{product} and 665 samples per \texttt{hazard}.


\section{Methods}
We used   ML and DL and implemented multiple data augmentation strategies. The next sections describe this in more detail.
\subsection{Machine Learning}

We used Term Frequency-Inverse Document Frequency (TF-IDF) \cite{SPARCKJONES1972} representation of text as input to our ML classifiers. We trained different classifiers and evaluated their performance on both subtasks for each category. The classifiers used were Linear Support Vector Machine (SVM), Decision Tree (DT), Random Forest (RF), Logistic Regression (LR), Multinomial Naive Bayes (NB), and K-Nearest Neighbors (KNN). We used the implementation from Scikit-learn library\footnote{\url{https://scikit-learn.org/stable/}}.

\subsection{Deep learning}
We used deep learning-based transformer language models for sequence classification \cite{vaswani2023attentionneed}. We chose encoder-only models that directly produce an input sequence's representation, which is fed into a classification head to make predictions. We trained various transformers for a sequence classification task, including BERT \cite{devlin-etal-2019-bert}, RoBERTa \cite{roberta}, DistilBERT \cite{distilbert}, and ModernBERT \cite{modernbert} (see \autoref{appendix:llm-details} for more details). We leveraged the Hugging Face's Transformers library\footnote{\url{https://huggingface.co/docs/transformers}} \cite{wolf-etal-2020-transformers}.

\subsection{Data augmentation on minority classes}
In addition to baseline training of the aforementioned models, we explored how data augmentation affected the performance of minority classes for each category. 

We employed three different augmentation strategies using the NLP AUG library\footnote{\url{https://nlpaug.readthedocs.io/}} \cite{ma2019nlpaug}: random word swapping (RW), synonym replacement (SR), and insertion of contextual words (CW). RW swapping randomly swaps adjacent words. SR substitutes similar words from a lexical database for the English language (WordNet \cite{wordnet}). CW uses contextual word embeddings from BERT to find the top similar words and insert them for augmentation.  An example of each technique applied to a \texttt{title} is shown in \autoref{tab:example:augmentation}.

\begin{table}
    \centering
    \resizebox{\columnwidth}{!}{%
    \begin{tabular}{c|p{7cm}}
    \hline
         \textbf{Operation}&\textbf{Sentence}\\\hline
        Original& Certain Stella Artois brand Beer may be unsafe due to possible presence of glass particles\\\hline
         CW& certain \textbf{notable} stella \textbf{by} artois brand beer may be \textbf{judged} unsafe \textbf{primarily} due to \textbf{his} possible presence of glass particles\\
         SR& Certain \textbf{Frank stella} Artois brand Beer may be \textbf{insecure} \textbf{imputable} to \textbf{potential} presence of glass particles\\
         RW& Certain Stella Artois brand Beer may \textbf{due} \textbf{be} unsafe to \textbf{presence} \textbf{possible} of glass particles\\\hline
    \end{tabular}
    }
    \caption{Examples of text augmentation techniques applied to a \texttt{title} of a food recall using contextual word insertion (CW), synonym replacement (SR), and random word swapping (RW).}
    \label{tab:example:augmentation}
        \vspace{-0.5cm}
\end{table}

For each strategy, we generated new samples in the training data for minority classes per category by altering titles and texts to preserve their inherent meaning while maintaining the annotated classes. For coarse categories (\texttt{hazard-category} and \texttt{product-category}), we augmented classes with fewer than 200 samples by generating 200 samples for each class. For fine-grained categories (\texttt{hazard} and \texttt{product}), we created 100 samples for classes with fewer than 100 samples for the \texttt{hazard} category and 50 samples for the \texttt{product} category.  After examining the entire class distributions, we chose these numbers of added samples and thresholds for low-support classes because they reflect a compromise between improving the representation of minority classes and maintaining low computational costs, but not completely resolving the imbalance issue. We first iterated through the existing data samples for each under-represented class of each category. We then distributed the specified total number of augmentation samples proportionally across these samples (adjusting the final one to ensure the addition matches the set target number of samples to add) to generate new samples based on the augmentation technique used. 
A pseudocode description is provided in \autoref{pseudocode} and its impact on class statistics is provided in \autoref{statistics-augmentation}. All methods were implemented in Python.

\section{Experiments}
 
In the next subsections, we further describe the preprocessing, hyperparameter fine-tuning, and evaluation details.

\subsection{Preprocessing}
Preprocessing included removal of HTML markup and special characters (newlines, tabs, Unicode character symbols) using regular expressions from \texttt{title} and \texttt{text} and text normalization such as whitespace standardization. This preserves semantic content while eliminating and filtering unnecessary formatting.

\subsection{Hyperparameter fine-tuning}
We fine-tuned the hyperparameters of baseline and augmented models on the \textbf{development} set using the Tree-structured Parzen Estimator (TPE) sampler in the Optuna hyperparameter optimization framework \citep{optuna_2019}. TPE is a Bayesian-based optimization approach that uses a tree structure to link between the hyperparameters and our objective function (maximizing $F_{1}$-macro score per category) to discover the optimal hyperparameters.

We ran ten trials per model and for each augmentation technique. For the ML for ST1, we ran 50 trials since the computation time was low. We optimized the parameters of the TF-IDF vectorizer, such as the minimum document frequency ($min\_df$), and hyperparameters applicable to each classifier for   ML, such as the maximum number of iterations ($max\_iter$) in SVM, and the learning rate scheduler, batch size, and epochs for DL (\autoref{hyperparameters}). 
All experiments involving transformer models were conducted on different GPU clusters (\autoref{cluster-details})\footnote{ Our best fine-tuned models are available at \url{https://huggingface.co/collections/DataScienceWFSR/semeval2025task9-food-hazard-detection-680f43d99cc294f617104be2}.}.

\subsection{Evaluation on leaderboard}
We submitted our results to the leaderboard for both subtasks, which calculated the final score by averaging the hazard $F_{1}$-macro (computed on all samples) with the product $F_{1}$-macro (computed only on samples with correct hazard predictions) for the coarse (ST1) and fine-grained categories (ST2). For example, if all hazards were predicted correctly, but all products were predicted incorrectly, the overall result would be a 0.5 $F_{1}$-macro score (\autoref{f1-metric}).


\section{Results}
The next subsections show quantitative results for each model in the official test set using the \texttt{text} field (trained in training and development sets) and an error analysis on the BERT baseline model versus its augmented-trained versions.

\begin{table}[t]
  \centering
\begin{adjustbox}{width=0.48\textwidth}
  \small
  \begin{tabular}{P{2.4cm}|p{1cm}p{0.9cm}p{0.9cm}p{0.9cm}|p{0.4cm}p{0.5cm}}
    \hline
     \textbf{Model} & \textbf{hazard-category} & \textbf{product-category} & \textbf{hazard} & \textbf{product}&\textbf{ST1}&\textbf{ST2} \\
    \hline
        $SVM_{base}$  & 0.701 & 0.626 & 0.544 & 0.234 & 0.682 & 0.396 \\
        $SVM_{CW}$& 0.655 & 0.642 & 0.519 & 0.256 & 0.649 & 0.396 \\ 
        $SVM_{SR}$ & 0.707 & 0.674 & 0.511 & 0.234 & 0.693 & 0.379 \\
        $SVM_{RW}$& 0.687 & 0.643 & 0.542 & 0.246 & 0.682 & 0.401 \\ \hline
        $LR_{base}$  & 0.666 & 0.665 & 0.511 & 0.203 & 0.680 & 0.368 \\ 
        $LR_{CW}$ & 0.713 & 0.682 & 0.457 & 0.209 & 0.702 & 0.347 \\ 
        $LR_{SR}$ & 0.698 & 0.677 & 0.454 & 0.233 & 0.691 & 0.354 \\
        $LR_{RW}$& 0.666 & 0.676 & 0.522 & 0.216 & 0.673 & 0.380 \\ \hline
        $DT_{base}$  & 0.542 & 0.445 & 0.405 & 0.012 & 0.484 & 0.208 \\ 
        $DT_{CW}$ & 0.617 & 0.491 & 0.427 & 0.029 & 0.544 & 0.230 \\ 
        $DT_{SR}$ & 0.576 & 0.488 & 0.464 & 0.037 & 0.526 & 0.252 \\ 
        $DT_{RW}$ & 0.612 & 0.475 & 0.506 & 0.056 & 0.542 & 0.283 \\ \hline
        $RF_{base}$  & 0.691 & 0.523 & 0.499 & 0.129 & 0.609 & 0.318 \\ 
       $RF_{CW}$ & 0.708 & 0.597 & 0.566 & 0.169 & 0.642 & 0.380 \\ 
        $RF_{SR}$ & 0.688 & 0.578 & 0.455 & 0.188 & 0.633 & 0.331 \\ 
       $RF_{RW}$ & 0.698 & 0.546 & 0.567 & 0.202 & 0.612 & 0.397 \\ \hline
        $KNN_{base}$  & 0.552 & 0.497 & 0.384 & 0.157 & 0.527 & 0.294 \\ 
        $KNN_{CW}$ & 0.565 & 0.490 & 0.376 & 0.169 & 0.534 & 0.309 \\ 
         $KNN_{SR}$& 0.552 & 0.507 & 0.389 & 0.163 & 0.537 & 0.305 \\ 
         $KNN_{RW}$ & 0.500 & 0.491 & 0.397 & 0.152 & 0.515 & 0.299 \\ \hline
        $NB_{base}$  & 0.553 & 0.570 & 0.306 & 0.064 & 0.568 & 0.203 \\ 
        $NB_{CW}$ & 0.599 & 0.586 & 0.405 & 0.175 & 0.603 & 0.310 \\ 
        $NB_{SR}$ & 0.588 & 0.574 & 0.444 & 0.140 & 0.589 & 0.314 \\ 
         $NB_{RW}$ & 0.603 & 0.617 & 0.383 & 0.167 & 0.631 & 0.300 \\ \hline
    \hline
    $BERT_{base}$ & 0.747 & 0.757 &	0.581 & 0.170 &	0.753 &	0.382\\
    $BERT_{CW}$  & 0.760 &	\textbf{0.761} &	\textbf{0.671} &	\textbf{0.280} &	0.762 &	\textbf{0.491}\\
    $BERT_{SR}$  & 0.770 &	0.754 &	0.666 &	0.275 &	0.764 &	0.478\\
    $BERT_{RW}$  & 0.752 &	0.757 &	0.651 &	0.275 &	0.756 &	0.467\\
    \hline
    $DistilBERT_{base}$  & 0.761 & 0.757	& 0.593 & 0.154 & 0.760 & 0.378\\
    $DistilBERT_{CW}$  & 0.766 &	0.753 &	0.635 &	0.246 &	0.763 &	0.449\\
     $DistilBERT_{SR}$ & 0.756 &	0.759 &	0.644 &	0.240 &	0.763 &	0.448\\
     $DistilBERT_{RW}$ &0.749 & 0.747 & 0.647 & 0.261 & 0.753 & 0.462\\
    \hline
    $RoBERTa_{base}$ & 0.760 & 0.753 & 0.579 & 0.123 & 0.755 & 0.356 \\
    $RoBERTa_{CW}$  &0.773 & 0.739 & 0.630 & 0.000 & 0.760 & 0.315\\
    $RoBERTa_{SR}$ &0.777 & 0.755 & 0.637 & 0.000 & 0.767 & 0.319\\
    $RoBERTa_{RW}$ &0.757 & 0.611 & 0.615 & 0.000 & 0.686 & 0.308\\
    \hline
    $ModernBERT_{base}$  &0.781	&0.745	&0.667	&0.275	&\textbf{0.769}	&0.485\\
    $ModernBERT_{CW}$  & 0.761	&0.712	&0.609	&0.252	&0.741	&0.441 \\
    $ModernBERT_{SR}$  &\textbf{0.790}	&0.728	&0.591	&0.253&	0.761	&0.434 \\
   $ModernBERT_{RW}$ & 0.761	&0.751	&0.629	&0.237	&0.759	&0.440 \\
    \hline
  \end{tabular}
\end{adjustbox}
\caption{$F_{1}$-macro scores for each model in the official test set given by the organizers utilizing the \texttt{text} field per category and subtasks scores (ST1 and ST2) rounded to 3 decimals. With bold, we indicated the higher score per category and subtask score.}
 \label{tab:results-text}
 \vspace{-0.6cm}
\end{table}

\subsection{Quantitative results}
\label{results}
Transformer models outperformed ML across all categories, as shown in \autoref{tab:results-text}, with the $ModernBERT_{base}$ leading across transformer models, in the \textbf{baseline} version, in all categories except \texttt{product-category}.

Among ML, SVM, LR, and RF showed competitive performance: $LR_{CW}$ scored highest in \texttt{hazard-category} (0.713) and \texttt{product-category} (0.682); $RF_{RW}$ in \texttt{hazard} (0.567), and $SVM_{CW}$ in \texttt{product} (0.256). 
Among the transformer models, $ModernBERT_{SR}$ scored highest in the \texttt{hazard-category} with a score of 0.790, while $BERT_{CW}$ scored highest in other categories. Augmentation increased performance but was not consistent across the categories. It was more pronounced in ST2 categories than ST1 categories, with the largest score increase (0.11) between $BERT_{base}$ and $BERT_{CW}$ augmentation in the \texttt{product} category. 

To understand the impact of augmentation, we conducted individual pairwise Kruskal-Wallis tests comparing the $F_1$-macro scores on the $BERT_{base}$ model with the augmented versions, training each version three times per category (\autoref{tab:model_stats}). Statistical significance (P < 0.05) was found in \texttt{product-category} with RW, in \texttt{hazard} with all augmentation techniques, and \texttt{product} with CW and RW (\autoref{tab:pvalues}). This indicates that augmentation techniques for BERT enhanced performance in minority classes more effectively in fine-grained categories than in coarse categories.

We submitted a combination of BERT and RoBERTa models for each category to the leaderboard (\autoref{official-models}), which resulted in an $F_1$-macro score of 0.761 for ST1 and of 0.453 for ST2 in the test set. These models were chosen since they indicated the best $F_1$-macro scores on the development set.
The other models were also evaluated on the test set, but not included in the leaderboard. The best scores achieved on ST1 was 0.769 and ST2 was 0.491, indicated in bold in \autoref{tab:results-text}. 

Moreover, experiments using only \texttt{title} were conducted (where their results can be found in \autoref{tab:results-title}). We continue with the error analysis on the models using \texttt{text} field since we observed better performance.
\begin{table}[h!]
\centering
\scalebox{0.8}{
\begin{tabular}{lccc}
\hline
\textbf{Category} & \textbf{CW} & \textbf{RW} & \textbf{SR} \\ 
\hline
hazard-category & 0.5127 & 0.2752 & 0.2752 \\
product-category & 0.2752 & 0.3758 & \textbf{0.0463} \\
hazard & \textbf{0.0495} & \textbf{0.0495} & \textbf{0.0463} \\
product & \textbf{0.0463} &\textbf{0.0495} & 0.5127 \\
\hline
\end{tabular}
}
\caption{Raw P-values from individual pairwise Kruskal-Wallis tests between $BERT_{base}$ model and each of the three augmentation techniques (rounded up to 4 decimals).}

\label{tab:pvalues}
\vspace{-0.7cm}
\end{table}

\subsection{Error Analysis - Confusion Matrices}

 We investigated the performance and shortcomings on the BERT model, which improved most with the CW technique compared to the baseline. 
 
\begin{figure}[t]
    \centering
    \includegraphics[width=1\linewidth]{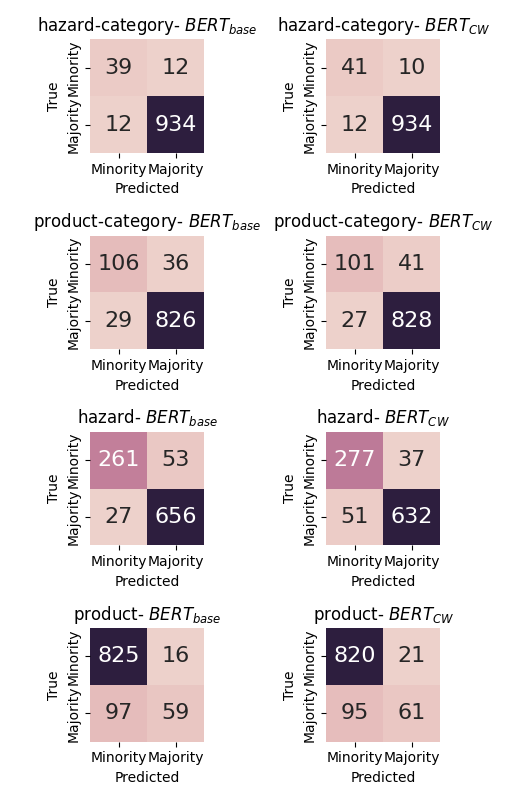}
    \caption{Confusion matrices comparing the performance of the $BERT_{base}$ and $BERT_{CW}$  models in the test set across the four categories showing the changes in the model's performance for minority and majority class predictions.}
    \label{fig:confusion-matrix}
    \vspace{-0.5cm}
\end{figure}

 When comparing the majority and minority classes that were augmented, the $BERT_{CW}$ model predicted the minority classes slightly better than $BERT_{base}$, with a rise from 39 to 41 for \texttt{hazard-category} and from 261 to 277 for \texttt{hazard} (around 6\% increase) (\autoref{fig:confusion-matrix}). However, the model predicted the majority classes slightly worse, decreasing from 656 to 632 for \texttt{hazard}, showing that there is a trade-off between improving the predictions for the minority versus the majority classes. Additionally, while the augmentation slightly improved the prediction of majority classes for the \texttt{product-category}, it decreased the minority class predictions from 106 to 101 samples. 

\begin{figure*}[ht]
\centering
\begin{subfigure}{0.45\textwidth}
  \centering
  \includegraphics[width=\linewidth]{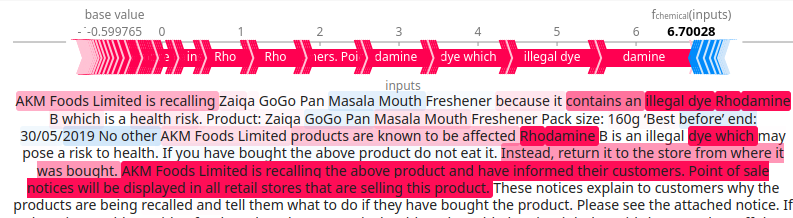}
  \caption{Visualization of a correctly classified sample for \texttt{hazard-category} in baseline model for the true \texttt{chemical} class.}
  \label{fig:sub1}
\end{subfigure}%
\hfill
\begin{subfigure}{0.45\textwidth}
  \centering
  \includegraphics[width=\linewidth]{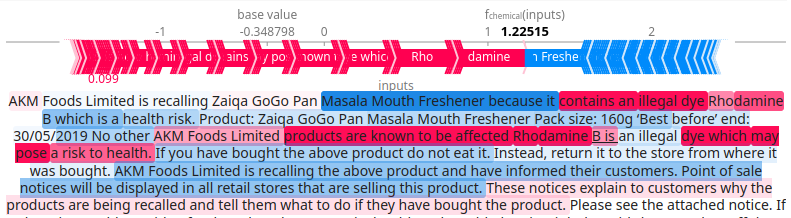}
  \caption{Visualization of a wrongly classified sample for \texttt{hazard-category} in CW augmentation model for the true \texttt{chemical} class.}
  \label{fig:sub2}
\end{subfigure}

\vspace{0.1cm}

\begin{subfigure}{0.45\textwidth}
  \centering
  \includegraphics[width=\linewidth]{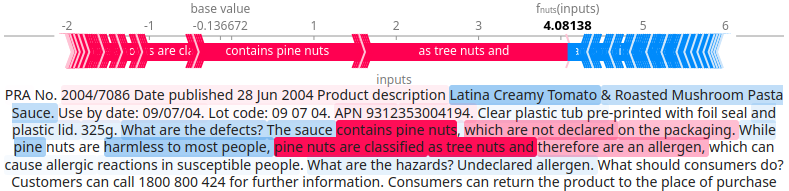}
  \caption{Visualization of a wrongly classified sample for \texttt{hazard} in baseline model for the true \texttt{nuts} class.}
  \label{fig:sub3}
\end{subfigure}%
\hfill
\begin{subfigure}{0.45\textwidth}
  \centering
  \includegraphics[width=\linewidth]{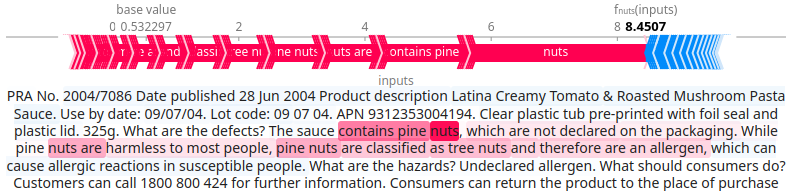}
  \caption{Visualization of a correctly classified sample for \texttt{hazard} in CW augmentation model for the true \texttt{nuts} class.}
  \label{fig:sub4}
\end{subfigure}

\caption{Visualizations of SHAP values for samples in the ground-truth classes of \texttt{hazard-category} and \texttt{hazard}, based on predictions from the $BERT_{base}$ and  $BERT_{CW}$ models. Pink text indicates positive contributions toward predicting this class, while blue text indicates negative contributions. Text which does not contribute in each sample has been truncated.}
\label{fig:images}
\vspace{-0.5cm}

\end{figure*}

\subsection{Error Analysis - SHAP}
We use SHapley Additive exPlanations (SHAP)\footnote{\url{https://shap.readthedocs.io/en/latest/}} to further analyze BERT’s prediction behavior. Figures \ref{fig:sub1} and \ref{fig:sub2} illustrate the SHAP values for a sample that was correctly classified with $BERT_{base}$, but was misclassified with $BERT_{CW}$ in the \texttt{hazard-category}, visualizing the contributions for the correct \texttt{chemical} class. Figures \ref{fig:sub3} and \ref{fig:sub4} show the SHAP values for a sample misclassified with $BERT_{base}$ but correctly classified with $BERT_{CW}$ in \texttt{hazard}, visualizing the contributions for the correct \texttt{nuts} class. For the \texttt{hazard-category}, 
the $BERT_{base}$ correctly identifies features such as `illegal dye' (in pink color), while the CW augmentation has more negative (blue) contributions that push the model's prediction away from the correct class. For the \texttt{hazard} category, although both models focus on significant terms like `pine nuts', the baseline model focuses on negative contributions like `Latina Creamy Tomato' resulting in a misclassification which may imply that the model associates these features incorrectly with different hazards. This misclassification pattern could serve as a basis for future investigation, further exploring and explaining the model's predictions to improve its performance and reliability.




\section{Limitations}
While multiple experiments have been conducted, some limitations could be addressed in future studies. The dataset used was exclusively in English, and the augmentation techniques applied were limited to word-level adjustments. Future research could explore more sophisticated augmentation methods, such as LLMs, to generate new samples and verify their quality.  Incorporating datasets in other languages could provide insight into the effectiveness of augmentation techniques. Further investigation could also focus on optimizing the number of augmented samples for minority classes to enhance classification performance, especially for food hazard classification, where reliable models are required to ensure safety. Lastly, to enhance even further the classifiers' performance, more complex architectures such as ensemble or hierarchical approaches could be used to compare their effectiveness on augmentation in the food hazard classification task. 

\section{Conclusion}
We showed that word-level text augmentation can enhance multi-class classification in minority classes. We used various machine learning and transformer models on the SemEval-2025 Task 9 to assess the effects of these augmentations. Leveraging the \texttt{text} field, we discovered that transformers tend to outperform ML. Augmentation techniques showed a slight increase of $F_{1}$-macro scores, but this effect was not consistent across all augmentations. Comparing $BERT_{base}$ with each augmentation technique, a statistical significant improvement was found for fine-grained categories, which indicates that augmenting minority classes can improve the performance of transformers for these classes.

\section*{Acknowledgments}
We thank anonymous reviewers for their feedback and the organizers for their support. Funding for this research has been provided by the European Union’s Horizon Europe research and innovation programme EFRA [grant number 101093026].

\bibliography{anthology,custom}

\appendix

\section{Dataset and Experiments Details}
\label{sec:appendix}
\subsection{Dataset Details}
\begin{figure*}[h!]
    \centering
    \includegraphics[width=0.8\linewidth]{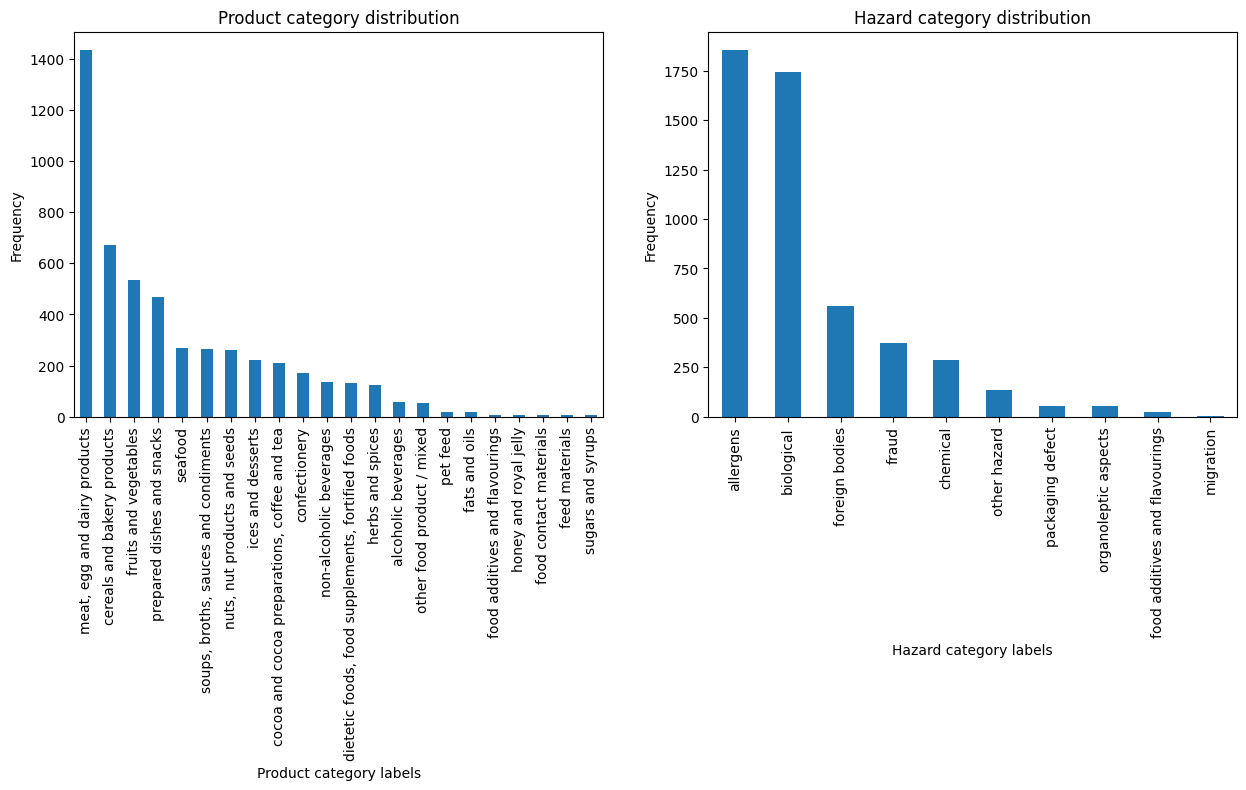}
    \caption{Distributions of \texttt{hazard-category} and \texttt{product-category} for classes occurrences.}
    \label{fig:dist_cat}
\end{figure*}

\begin{figure*}[h!]
    \centering
    \includegraphics[width=0.8\linewidth]{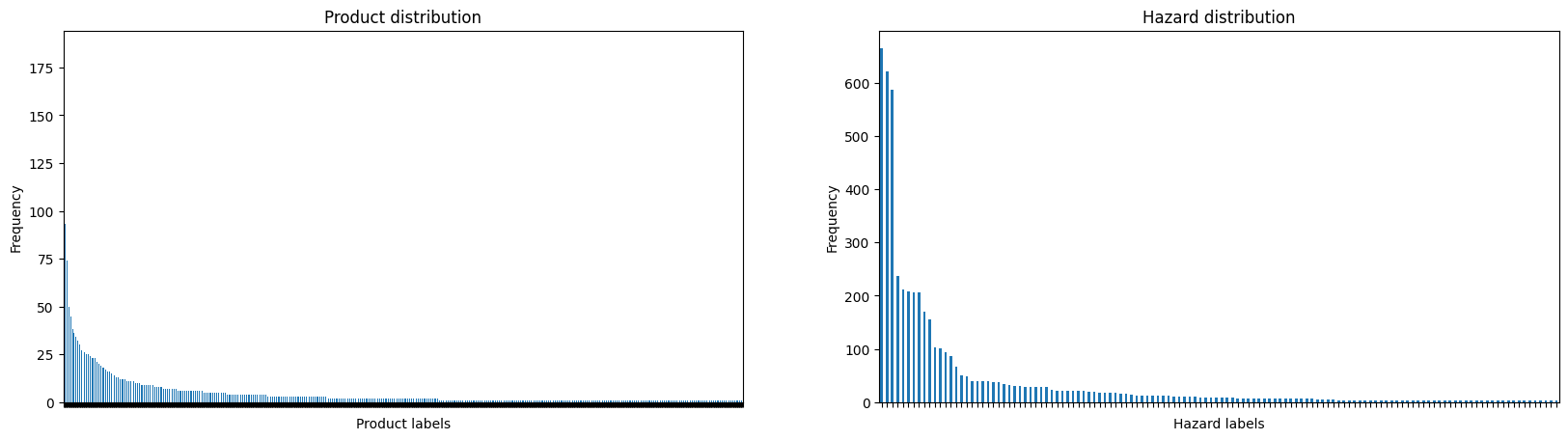}
    \caption{Distributions of \texttt{hazard} and \texttt{product} for classes occurrences. The classes in the x-axis have been omitted due to the large number of classes and clearness of the chart.}
    \label{fig:dist_}
\end{figure*}

\begin{figure*}[h!]
    \centering
    \includegraphics[width=0.8\linewidth]{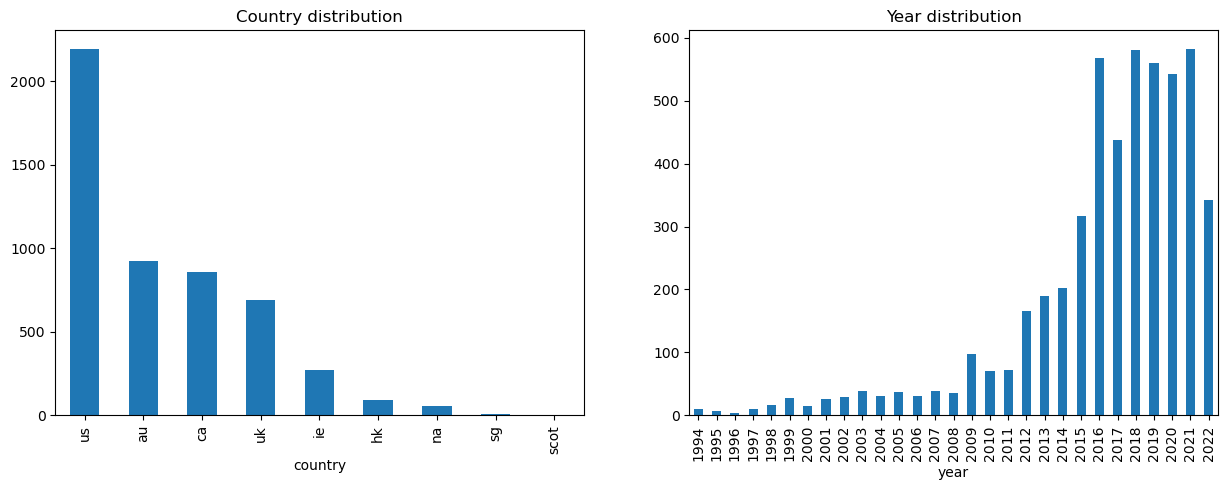}
    \caption{Distributions of occurrences per country (left figure) and per year (right figure) published in the given dataset.}
    \label{fig:distribution}
\end{figure*}

In this section, tables and figures related to the statistics of the provided dataset are presented. \autoref{tab:examples} shows some sample titles and text from the dataset along with their annotated classes. \autoref{tab:labels} presents the number and the names of the annotated classes. \autoref{fig:dist_cat} and \autoref{fig:dist_} show the distributions of hazard and product classes in coarse and fine-grained categories indicating the long-tail distributions that follow, while \autoref{fig:distribution} presents the distribution of occurrences in the dataset per country and year. \autoref{tab:domain_samples} displays the site domain that the samples have been sourced along their number of samples.

\begin{table}[h!]
\centering
\small
\begin{tabular}{|l|r|}
\hline
\textbf{Domain} & \textbf{Samples} \\
\hline
www.fda.gov & 1740 \\
www.fsis.usda.gov & 1112 \\
www.productsafety.gov.au & 925 \\
www.food.gov.uk & 902 \\
www.lebensmittelwarnung.de & 886 \\
www.inspection.gc.ca & 864 \\
www.fsai.ie & 358 \\
www.foodstandards.gov.au & 281 \\
inspection.canada.ca & 124 \\
www.cfs.gov.hk & 123 \\
recalls-rappels.canada.ca & 96 \\
tna.europarchive.org & 52 \\
wayback.archive-it.org & 23 \\
healthycanadians.gc.ca & 18 \\
www.sfa.gov.sg & 11 \\
www.collectionscanada.gc.ca & 8 \\
securite-alimentaire.public.lu & 6 \\
portal.efet.gr & 4 \\
www.foodstandards.gov.scot & 3 \\
www.ages.at & 2 \\
www.accessdata.fda.gov & 1 \\
webarchive.nationalarchives.gov.uk & 1 \\
www.salute.gov.it & 1 \\
www.foedevarestyrelsen.dk & 1 \\
\hline
\end{tabular}
\caption{Data sources of public food safety authority websites, ordered by support number of the given dataset. Table adapted from \citet{randl_2024_10891602}. It contains also the sources for the non-English data.}
\label{tab:domain_samples}
\end{table}

\subsection{Preprocessing Dataset Details}
The \texttt{html.parser} was leveraged using the BeautifulSoup\footnote{\url{https://www.crummy.com/software/BeautifulSoup/bs4/doc/}} package to remove the HTML content from the data.
The regular expression that was used to remove the special characters is the following: 
\begin{verbatim}
'[\t\n\r\u200b]|//|&nbsp'
\end{verbatim}

\subsection{System Configurations Details}
\label{cluster-details}
The experiments were run on different machines using Python version 3.10.16. For the fine-tuning and training of transformer models,  NVIDIA A100 80GB and NVIDIA GeForce RTX 3070 Ti were utilized. For reproducibility, we used $seed=2025$ as a seed number by employing it in PyTorch, NumPy, and Random packages. To run the BERT model two extra times and calculate the statistical significance, we used $seed = 2024$ and $2026$. Moreover, the package versions and their respective URLs that were leveraged can be found in \autoref{tab:python_libraries}.

\begin{table}[h]
    \centering
    \small
    \begin{tabular}{|l|l|p{3cm}|}
        \hline
        \textbf{Library} & \textbf{Version} & \textbf{URL} \\ \hline
        Transformers    &          4.49.0 & \url{https://huggingface.co/docs/transformers/index}\\
        PyTorch       &           2.6.0 &\url{https://pytorch.org/}\\
        SpaCy           &          3.8.4&\url{https://spacy.io/}\\
        Scikit-learn     &         1.6.0&\url{https://scikit-learn.org/stable/}\\
        Pandas           &         2.2.3&\url{https://pandas.pydata.org/}\\
        Optuna          &          4.2.1&\url{https://optuna.org/}\\
        NumPy          &           2.0.2&\url{https://numpy.org/}\\
        NLP AUG          &          1.1.11& \url{https://nlpaug.readthedocs.io/en/latest/index.html}\\
        BeautifulSoup4   &         4.12.3&\url{https://www.crummy.com/software/BeautifulSoup/bs4/doc/#}\\\hline
    \end{tabular}
    \caption{Python libraries and their versions with URLs used for the code implementation of the paper.}
    \label{tab:python_libraries}
\end{table}

\subsection{Transformer Models Details}
\label{appendix:llm-details}
In this section, we explain the encoder-only transformer models' details and architectures we used in the experiments. For BERT \cite{devlin-etal-2019-bert}, the \texttt{bert-base-uncased}\footnote{\url{https://huggingface.co/google-bert/bert-base-uncased}} is used which consists of 110M parameters, 12 encoder layers, a hidden state size of 768, a feed-forward hidden state of 3072, and 12 attention heads, serving as a foundational pre-trained transformer model. For RoBERTa \cite{roberta}, the \texttt{roberta-base}\footnote{\url{https://huggingface.co/FacebookAI/roberta-base}} (case-sensitive) is leveraged and has 125M parameters, structured with 12 encoder layers, a 768-dimensional hidden state, a 3072-dimensional feed-forward network, and 12 attention heads, which is trained on a large corpus leveraging dynamic masking. For DistilBERT \cite{distilbert}, the \texttt{distilbert-base-uncased}\footnote{\url{https://huggingface.co/distilbert/distilbert-base-uncased}} is utilized, which is a lighter BERT variant having 66M parameters and 6 encoder layers while maintaining similar hidden state size and attention heads with BERT. For the ModernBERT \cite{modernbert}, we used the \texttt{ModernBERT-base} \footnote{\url{https://huggingface.co/answerdotai/ModernBERT-base}} (case-sensitive), which contains 149M parameters, 22 encoder layers, hidden state of 768, an intermedia size of 1152, and 12 attention heads. It is trained on 2 trillion tokens, extending the token length to 8192 and incorporating other architectural enhancements to make it faster, lighter, and with better performance than other BERT variants.

\subsection{Text Augmentation Details}
\label{pseudocode}
In Algorithm \ref{algo:pseudocode}, the function for creating new samples using augmentation is presented. Starting from the inputs of the function, it accepts: a threshold $\tau$ which is the number of samples that a class could contain to be a minority class, the number of samples to add $S$ per minority class, a class counts $C$ that contains the number of samples per class, an augmentation function $F$ that accepts the sample and the number of samples to create, the original training dataset $D$ and the $category$ (e.g. \texttt{hazard}) that we want to augment its classes. The function begins with finding the minority classes by getting the classes with samples less than the given threshold. Then, for each minority class, the respective samples are collected and the number of samples that need to be augmented for each sample is calculated by dividing the total samples over the number of samples of the specific class rounding down the result to the nearest integer. For each sample, then the augmentation function is applied and creates new samples, except for the last sample which is augmented for the remaining number of samples needed. The new samples are inserted into the original training dataset and the function returns the augmented set.
\begin{algorithm}
\caption{Function of creating samples for classification with augmentation}
\small
\begin{algorithmic}[1]
\Require Threshold of class number of samples $\tau$, Total samples to add $S$, Class counts $C$, Augmentation function $F$ , Original training dataset $D$, Category to augment $cat$
\Function{create\_augmented\_samples}{$\tau$, $S$, $C$, $F$, $D$, $cat$}
\State $minority\_classes \gets \{c \mid C[c] < \tau\}$
\State $a\_s \gets \emptyset$ \Comment{augmented\_samples}
\For{$c$ in $minority\_classes$}
\State $samples \gets \{d \in D \mid d[cat] = c\}$
\State $N \gets \left\lfloor \frac{S}{|\text{samples}|} \right\rfloor $  
\For{$sample$ in $samples$}
\If{is the last $sample$}
\State $N \gets  S - [N  * (|samples|-1)]$
\EndIf
\State $new\_samples\ \gets F(sample, N) $
\State $a\_s \gets new\_samples \cup \{a\_s\}$
\EndFor
\EndFor
\State $augmented\_set \gets D \cup a\_s$

\Return $augmented\_set$
\EndFunction
\end{algorithmic}
\label{algo:pseudocode}
\end{algorithm}

\subsection{Dataset Classes Statistics}
\label{statistics-augmentation}
In \autoref{tab:statistics-augmentation}, a comparison between the classes' statistics before and after applying augmentation per category is presented. For \texttt{hazard-category} and \texttt{product-category}, the number of samples that have been created are 200 for classes that have under 200 samples. For \texttt{hazard} and \texttt{product}, the number of samples that have been added are 100 and 50, respectively, for classes that have under 100 samples.
\begin{table}[h!]

\centering
\resizebox{\columnwidth}{!}{%
\begin{tabular}{@{}l|cccc@{}}
\toprule
\textbf{Statistic}          & \multicolumn{2}{c}{\textbf{hazard-category}} & \multicolumn{2}{c}{\textbf{product-category}} \\
                   & Initial         & Augmented        & Initial         & Augmented         \\ \midrule
Count              & \multicolumn{2}{c}{10}             & \multicolumn{2}{c}{22}               \\
Mean               & 508.2           & 608.2            & 231.0           & 349.2             \\
Standard Deviation & 702.75          & 635.57           & 325.83          & 270.79            \\
Minimum            & 3               & 203              & 5               & 205               \\
25\%               & 53.25           & 253.25           & 19.25           & 212.25            \\
50\%               & 210.5           & 310.5            & 132.5           & 260.5             \\
75\%               & \multicolumn{2}{c}{513.5}          & 263.5           & 333.25            \\
Maximum            & \multicolumn{2}{c}{1854}           & \multicolumn{2}{c}{1434}             \\ 
Total Samples & 5082&6082 &5082&7682\\\midrule
\end{tabular}
}
\resizebox{\columnwidth}{!}{%
\begin{tabular}{@{}l|cccc@{}}
\toprule
\textbf{Statistic}          & \multicolumn{2}{c}{\textbf{hazard}}         & \multicolumn{2}{c}{\textbf{product}}          \\
                   & Initial         & Augmented        & Initial         & Augmented         \\ \midrule
Count              & \multicolumn{2}{c}{128}            & \multicolumn{2}{c}{1022}             \\
Mean               & 39.7            & 130.33           & 4.97            & 54.87             \\
Standard Deviation & 102.19          & 81.14            & 10.97           & 9.72              \\
Minimum            & 3               & 101              & 1               & 51                \\
25\%               & 4               & 104              & 1 & 51            \\
50\%               & 8.5             & 108              & 2               & 52                \\
75\%               & 24.25           & 122              & 4 & 54               \\
Maximum            & \multicolumn{2}{c}{665}            & \multicolumn{2}{c}{185}              \\ 
Total Samples & 5082&16682 &5082&56082\\
\bottomrule
\end{tabular}
}
\caption{Comparison between the initial and after augmentation classes' statistics per category (\texttt{hazard-category}, \texttt{product-category}, \texttt{hazard}, \texttt{product}) in the training dataset.}
\label{tab:statistics-augmentation}
\end{table}

\subsection{$F_{1}$ Macro Evaluation Metric}
\label{f1-metric}
For both subtasks, the evaluation metric given by the organizers was the $F_{1}$-macro score on the predicted and the annotated classes. The rankings are based on the hazard classes, meaning that if predictions for both hazard and product are correct, it will get a 1.0 score, while if the hazard predictions are correct but for product are wrong, it will score 0.5. The accurate scoring function can be seen in Algorithm \ref{algo:scoring}.
\begin{algorithm}
\small
\caption{Function for computing score for each subtask.}
\label{algo:scoring}
\begin{algorithmic}[1]
\Require hazard true $ht$, product true $pt$, hazard predictions $hp$, product predictions $pp$
\Function{compute\_score}{$ht$, $pt$, $hp$, $pp$}
    \State $\textit{$F_{1}$\_hazards} \gets \text{$F_{1}$-macro}(ht, hp)$
    \State $\textit{cm} \gets (hp == ht)$  \Comment{correct\_mask}
    \State $\textit{$F_{1}$\_products} \gets \text{$F_{1}$-macro}( pt[\textit{cm}], pp[\textit{ccm}])$
    
    \State \Return $\frac{1}{2}(\textit{$F_{1}$\_hazards} + \textit{$F_{1}$\_products})$
\EndFunction
\end{algorithmic}
\end{algorithm}

\subsection{Hyperparameters Details}
\label{hyperparameters}

To tune the hyperparameters, the Optuna optimization framework was employed, optimizing based on $F_{1}$-macro scores. For the ML models, the TF-IDF vectorizer parameters, such as $min_{df}$, $max\_df$ etc., were optimized, along with specific parameters for each model, such as $max\_iter$ for SVM and LR, and $alpha$ for NB. The utilized hyperparameters for each model, category, and field are presented in \cref{tab:hyperparameters:svm,tab:hyperparameters:lr,tab:hyperparameters:dt,tab:hyperparameters:rf,tab:hyperparameters:knn,tab:hyperparameters:nb}. When the SpaCy tokenizer\footnote{\url{https://spacy.io/api/tokenizer}} is used, English stopwords from SpaCy are also removed from the given text. Balanced class weight was used in SVM, LR, RF, and DT models.

For the transformer models, $batch\_size$, $epochs$, and $lr\_scheduler$ were optimized across all model variants over 10 trials. For all models, the learning rate was set at $5.0\mathrm{e} {-5}$, and the maximum token length that the tokenizer can generate was set at $128$, as no significant differences in performance with higher maximum token length were observed. In \cref{tab:hyperparameters:bert,tab:hyperparameters:roberta,tab:hyperparameters:distilbert,tab:hyperparameters:modernbert}, the utilized hyperparameters for each model, category, and field are listed.

The search space for each hyperparameter used during the tuning can be found in \autoref{tab:hyperparameters:search_space}.
\begin{table}[h]
\centering
\small
\begin{tabular}{lp{4.5cm}}
\toprule
\textbf{Hyperparameter} & \textbf{Search Space} \\
\midrule
$C$ & \{0.1, 1, 5, 10\} \\
$max\_iter$ & \{100, 1000, 5000\} \\
$n\_estimators$ & \{100, 200, 300\} \\
$max\_depth$ \textit{(DT)} & \{100, 200, 300\} \\
$max\_depth$ \textit{(RF)} & \{100, 1000, 5000\} \\
$max\_features$ & \{1000, 5000, 10000, 50000\} \\
$n\_neighbors$ & \{3, 5, 7, 9, 11\}\\
$weights$ & \{uniform, distance\}\\
$alpha$ &\{0.01, 0.1, 1, 5\}\\ 
$analyzer$ & \{word, char\} \\
$tokenizer$ & \{-, SpaCy\} \\
$min\_df$ & \{1, 2, 5\} \\
$max\_df$ & \{0.1, 0.3, 0.5\} \\
$ngram\_range$ & \{(1, 1), (1, 2), (1, 3), (1, 4), (1, 5), (2, 3), (2, 4), (2, 5), (3, 5)\} \\ \hline
$batch\_size$ & \{8, 16, 32\}\\
$epochs$ & \{3, 5, 10\}\\
$lr\_scheduler$ & \{lin, cos, cosRestarts\}\\
\bottomrule
\end{tabular}
\caption{Search space for each hyperparameter used in Optuna optimization trials for ML and transformer models. For learning rate schedulers: cos (cosine annealing), cosRestarts (cosine annealing with restarts), and lin (linear). }
\label{tab:hyperparameters:search_space}
\end{table}
\section{More Results and Explainability Analysis}

\subsection{Results using title}
\begin{table}[h!]
  \centering
\begin{adjustbox}{width=0.46\textwidth}
  \small
  \begin{tabular}{P{2.4cm}|p{1cm}p{0.9cm}p{0.9cm}p{0.9cm}|p{0.4cm}p{0.5cm}}
    \hline
     \textbf{Model} & \textbf{hazard-category} & \textbf{product-category} & \textbf{hazard} & \textbf{product}&\textbf{ST1}&\textbf{ST2} \\
    \hline
$SVM_{base}$& 0.644 & 0.692 & 0.436 & 0.250 & 0.670 & 0.363 \\  
$SVM_{CW}$& 0.641 & 0.675 & 0.402 & 0.240 & 0.657 & 0.343 \\  
$SVM_{SR}$& 0.646 & 0.699 & 0.435 & 0.259 & 0.674 & 0.364 \\  
$SVM_{RW}$&  0.646 & 0.690 & 0.432 & 0.253 & 0.670 & 0.372 \\ \hline  
$LR_{base}$& 0.596 & 0.695 & 0.419 & 0.261 & 0.636 & 0.359 \\  
$LR_{CW}$& 0.627 & 0.670 & 0.428 & 0.263 & 0.649 & 0.361 \\  
$LR_{SR}$& 0.612 & 0.660 & 0.425 & 0.234 & 0.639 & 0.350 \\  
$LR_{RW}$& 0.634 & 0.647 & 0.442 & 0.269 & 0.644 & 0.374 \\  \hline
$DT_{base}$& 0.491 & 0.478 & 0.330 & 0.036 & 0.483 & 0.183 \\  
$DT_{CW}$& 0.534 & 0.541 & 0.277 & 0.031 & 0.553 & 0.164 \\  
$DT_{SR}$& 0.565 & 0.449 & 0.349 & 0.081 & 0.495 & 0.226 \\  
$DT_{RW}$& 0.513 & 0.453 & 0.298 & 0.057 & 0.493 & 0.185 \\  \hline
$RF_{base}$& 0.611 & 0.633 & 0.420 & \textbf{0.287} & 0.616 & 0.369 \\  
$RF_{CW}$& 0.592 & 0.640 & 0.446 & 0.232 & 0.615 & 0.367 \\  
$RF_{SR}$& 0.638 & 0.527 & 0.422 & 0.207 & 0.590 & 0.329 \\  
$RF_{RW}$& 0.629 & 0.635 & 0.372 & 0.244 & 0.638 & 0.328 \\  \hline
$KNN_{base}$& 0.519 & 0.598 & 0.349 & 0.187 & 0.566 & 0.299 \\  
$KNN_{CW}$& 0.554 & 0.508 & 0.341 & 0.167 & 0.545 & 0.275 \\  
$KNN_{SR}$& 0.541 & 0.569 & 0.306 & 0.152 & 0.566 & 0.255 \\  
$KNN_{RW}$& 0.536 & 0.551 & 0.335 & 0.174 & 0.558 & 0.278 \\  \hline
$NB_{base}$& 0.597 & 0.641 & 0.366 & 0.221 & 0.624 & 0.318 \\  
$NB_{CW}$& 0.588 & 0.611 & 0.360 & 0.185 & 0.609 & 0.305 \\  
$NB_{SR}$& 0.597 & 0.593 & 0.349 & 0.180 & 0.600 & 0.290 \\  
$NB_{RW}$&0.585 & 0.629 & 0.390 & 0.195 & 0.608 & 0.315 \\
\hline \hline
$BERT_{base}$& 0.668 & 0.636 & 0.372 & 0.177 & 0.653 & 0.284 \\  
$BERT_{CW}$& 0.654 & 0.714 & 0.502 & 0.249 & 0.693 & 0.392 \\  
$BERT_{SR}$& 0.650 & 0.707 & 0.489 & 0.259 & 0.681 & 0.389 \\  
$BERT_{RW}$& \textbf{0.670} & 0.735 & 0.477 & 0.250 & \textbf{0.700} & 0.372 \\  \hline
$DistilBERT_{base}$ & 0.653 & 0.579 & 0.396 & 0.248 & 0.613 & 0.334 \\  
$DistilBERT_{CW}$ & 0.631 & 0.725 & 0.486 & 0.264 & 0.687 & 0.395 \\  
$DistilBERT_{SR}$ & 0.640 & 0.695 & \textbf{0.503} & 0.262 & 0.667 & \textbf{0.400} \\  
$DistilBERT_{RW}$ & 0.644 & 0.701 & 0.496 & 0.267 & 0.672 & 0.392 \\  \hline
$RoBERTa_{base}$& 0.608 & 0.629 & 0.384 & 0.076 & 0.619 & 0.246 \\  
$RoBERTa_{CW}$ & 0.668 & 0.692 & 0.460 & 0.000 & 0.686 & 0.230 \\  
$RoBERTa_{SR}$ & 0.639 & 0.718 & 0.471 & 0.000 & 0.673 & 0.236 \\  
$RoBERTa_{RW}$ & 0.636 & \textbf{0.736} & 0.479 & 0.001 & 0.690 & 0.240 \\  \hline
$ModernBERT_{base}$& 0.586 & 0.671 & 0.393 & 0.275 & 0.627 & 0.353 \\  
$ModernBERT_{CW}$& 0.649 & 0.731 & 0.423 & 0.266 & 0.688 & 0.372 \\  
$ModernBERT_{SR}$& 0.616 & 0.679 & 0.422 & 0.254 & 0.646 & 0.364 \\  
$ModernBERT_{RW}$& 0.641 & 0.697 & 0.385 & 0.263 & 0.668 & 0.351 \\  
\hline
\end{tabular}
\end{adjustbox}
\caption{$F_{1}$-macro scores in the official test set given by the organizers utilizing the \texttt{title} field per category and subtasks scores (ST1 and ST2) rounding up to 3 decimals. With bold, we indicate the higher score per column.}
\label{tab:results-title}
\end{table}
In \autoref{tab:results-title}, we present the experimental results on the test set using the \texttt{title} field for both ML and transformer models. As with the results using \texttt{text}, transformer models overall outperformed the ML models, although they were lower than using \texttt{text}. The best models per category are: $BERT_{RW}$ for \texttt{hazard-category} (0.670), $RoBERTa_{RW}$ for \texttt{product-category} (0.736), $DistilBERT_{SR}$ for \texttt{hazard} (0.503), and $RF_{base}$ for \texttt{product} (0.287). Among the ML models, SVM, LR, and RF demonstrated competitive performance across the categories, similar to the performance observed using only the \texttt{text} field. While there was variability between the baseline and augmented models, a slight, consistent increase was observed in \texttt{product-category} and \texttt{hazard} when using transformer models.

\subsection{Statistical Significance Experiments}
The mean $F_1$-macro scores for the BERT model experiments (both baseline and augmented versions, each run three times) are presented in \autoref{tab:model_stats}.
\begin{table}[h!]
\centering

\scalebox{0.9}{
\begin{tabular}{p{1.3cm}P{1.2cm}P{1.2cm}P{1.2cm}P{1.2cm}}
\hline
\textbf{Model} & \textbf{hazard-category} & \textbf{product-category} & \textbf{hazard} & \textbf{product}  \\ 
\hline
$BERT_{base}$ & $0.757$ & $0.769$ & $0.594$ & $0.186$ \\ 
$BERT_{CW}$ & $0.768$ & $0.756$ & $0.658$ & $0.284$ \\
$BERT_{RW}$  & $0.751$ & $0.752$ & $0.662$ & $0.256$ \\
$BERT_{SR}$  & $0.771$ & $0.75$ & $0.652$ & $0.189$ \\

\hline
\end{tabular}
}
\caption{Mean $F_{1}$-macro scores per category for each $BERT_{base}$ and with augmentation models running three times using as random seed numbers: 2024, 2025, and 2026.}
\label{tab:model_stats}
\end{table}
\subsection{Official Submitted Models}
\label{official-models}
Since only one submission was allowed during the evaluation phase, the predictions of the models that were submitted and were found to have the best $F_{1}$-macro scores on the development set for each category are: $RoBERTa_{base}$ for \texttt{hazard-category} with 0.880 $F_{1}$-macro score, $RoBERTa_{RW}$ for \texttt{product-category} with 0.750 $F_{1}$-macro score, $BERT_{CW}$ for \texttt{hazard} with 0.682 $F_{1}$-macro score, $BERT_{RW}$ for \texttt{product} with 0.260 $F_{1}$-macro score (all trained in \texttt{text} field). Then, these models were trained in both train and dev sets and provided their predictions on the test set. When submitting this combination of models, an ST1 score of 0.761 and an ST2 score of 0.4529 were achieved, which are our official leaderboard scores.

\begin{table*}[h!]
\scalebox{0.7}{
\centering
\begin{tabular}{@{}p{0.2\linewidth}p{0.1\linewidth}p{1.1\linewidth}@{}}
\toprule
\textbf{Category} & \textbf{Number of Classes} &  \textbf{Names of Classes} \\ \midrule
Hazard Category& 10& `allergens', `biological', `foreign bodies`, `fraud', 'chemical', `other hazard', `packaging defect', `organoleptic aspects', `food additives and flavourings', `migration' \\ 
Product Category& 22& `meat, egg and dairy products', `cereals and bakery products', `fruits and vegetables', `prepared dishes and snacks', `seafood', `soups, broths, sauces and condiments', `nuts, nut products and seeds',`ices and desserts',`cocoa and cocoa preparations', `coffee and tea',`confectionery',`non-alcoholic beverages',`dietetic foods', `food supplements', `fortified foods',`herbs and spices',`alcoholic beverages',`other food product / mixed',`pet feed',`fats and oils',`food additives and flavourings',`honey and royal jelly',`food contact materials', `feed materials', `sugars and syrups'\\ 
Hazard&128&`listeria monocytogenes',`salmonella',`milk and products thereof',`escherichia coli',`peanuts and products thereof' ... `dioxins',`staphylococcal enterotoxin',`dairy products',`sulfamethazine unauthorised',`paralytic shellfish poisoning (psp) toxins'\\
Product& 1068 & `ice cream', 'chicken based products', `cakes', `ready to eat - cook meals', `cookies' ... `breakfast cereals and products therefor', `dried lilies', `chilled pork ribs', `tortilla chips cheese', `ramen noodles' \\ \bottomrule
\end{tabular}
}
\caption{Names and number of total classes of the four annotated categories. For \texttt{hazard} and \texttt{product}, some classes are ommited. For \texttt{product}, the total number of classes along with the test data is 1,142.}
\label{tab:labels}
\end{table*}
\begin{table*}[h!]
\centering
\scalebox{0.7}{
\begin{tabular}{@{}p{0.2\linewidth}p{0.7\linewidth}p{0.1\linewidth}p{0.1\linewidth}p{0.1\linewidth}p{0.1\linewidth}@{}}
\toprule
\textbf{Title} & \textbf{Text} &  \textbf{hazard-category} & \textbf{hazard} & \textbf{product-category} & \textbf{product} \\ \midrule
 Wismettac Asian Foods Issues Allergy Alert on Undeclared Wheat and Soy in Dashi Soup Base& Wismettac Asian Foods, Inc., Santa Fe Springs, CA is recalling 17.6 oz packages of Marutomo Dashi Soup Base because they may contain undeclared wheat and soy. ... Consumers with questions may contact the company at recall@wismettacusa.com. & allergens& soybeans and products thereof & soups, broths, sauces and condiments & soups\\\hline
Kader Exports Recalls Frozen Cooked Shrimp Because of Possible Health Risk & Kader Exports, with an abundance of caution, is recalling certain consignments of various sizes of frozen cooked, peeled and deveined shrimp sold in 1lb, 1.5lb., and 2lb. retail bags.  ... Consumers with questions may contact the company at +91-022-62621004/ +91-022-62621009, Mon-Fri 10:00hrs -16:00hrs GMT+5.5. &biological & salmonella & seafood & shrimps\\ \hline
Recall Notification: FSIS-024-94 & Case Number: 024-94 Date Opened: 07/01/1994 ...       Product:  SMOKED CHICKEN SAUSAGE Problem:  BACTERIA  Description: LISTERIA Total Pounds Recalled:  2,894  Pounds Recovered:  2,894 &  biological & listeria monocytogenes & meat, egg and dairy products & smoked sausage\\ \bottomrule
\end{tabular}
}
\caption{Samples from the Food Recall Incidents dataset with title, text and the annotated categories.}
\label{tab:examples}
\end{table*}
\newpage



\begin{table*}[h]
\small
    \scalebox{0.82}{
    \begin{tabular}{lcccccccc}
        \multicolumn{9}{c}{\textbf{Hyperparameters for SVM}}\\
    \toprule[1.5pt]

& \multicolumn{4}{c}{\textbf{hazard-category}} & \multicolumn{4}{c}{\textbf{product-category}}\\
        \textbf{Parameters}& \textbf{baseline} & \textbf{CW} & \textbf{SR} & \textbf{RW}& \textbf{baseline} & \textbf{CW} & \textbf{SR} & \textbf{RW} \\
       & title / text & title / text & title / text & title / text& title / text & title / text & title / text & title / text \\\midrule
$C$ & 5 & 1 & 1 / 10 & 10 / 1 & 1 / 10 & 10 & 1 & 5 / 10 \\
$max\_iter$ & 1000 / 5000 & 5000 / 100 & 5000 & 1000 / 100 & 5000 / 100 & 1000 & 5000 & 100 / 5000 \\
$max\_features$ & 50000 & 50000 & 50000 / 10000 & 50000 / 5000 & 50000 & 50000 & 50000 & 50000 \\
$analyzer$ & char & word & word / char & char & char / word & char / word & char & char / word \\
$tokenizer$ & - & SpaCy / - & SpaCy / - & - & - & - & - & - \\
$max\_df$ & 0.5 / 0.3 & 0.1 & 0.5 / 0.3 & 0.5 & 0.5 / 0.1 & 0.1 / 0.5 & 0.1 & 0.3 / 0.5 \\
$min\_df$ & 1 / 5 & 1 / 2 & 1 & 1 / 2 & 1 / 2 & 5 & 2 / 1 & 5 / 2 \\
$ngram\_range$ & (2, 5) / (1, 5) & (1, 3) / (2, 4) & (1, 2) / (3, 5) & (2, 5) / (2, 4) & (1, 5) / (1, 4) & (2, 5) / (1, 3) & (3, 5) & (2, 5) / (1, 3) \\
 \bottomrule[1.5pt]
  & \multicolumn{4}{c}{\textbf{hazard}} & \multicolumn{4}{c}{\textbf{product}}\\
 \textbf{Parameters}& \textbf{baseline} & \textbf{CW} & \textbf{SR} & \textbf{RW}& \textbf{baseline} & \textbf{CW} & \textbf{SR} & \textbf{RW} \\   
& title / text & title / text & title / text & title / text& title / text & title / text & title / text & title / text \\\midrule
$C$ & 5 / 10 & 1 / 10 & 5 & 1 / 5 & 10 & 5 / 1 & 10 / 5 & 5 / 10 \\
$max\_iter$ & 1000 / 5000 & 5000 / 1000 & 1000 / 5000 & 1000 / 5000 & 100 / 1000 & 1000 / 100 & 1000 & 5000 / 1000 \\
$max\_features$ & 50000 & 10000 / 50000 & 50000 / 10000 & 50000 & 5000 & 5000 / 50000 & 50000 & 10000 / 50000 \\
$analyzer$  & char & char & word & word / char & char / word & char & char & char \\
$tokenizer$ & - & - & - & - & - / SpaCy & - & - & - \\
$max\_df$  & 0.5 / 0.1 & 0.5 / 0.1 & 0.1 / 0.3 & 0.3 / 0.5 & 0.1 / 0.5 & 0.1 & 0.3 & 0.5 / 0.1 \\
$min\_df$ & 5 / 1 & 1 / 2 & 2 / 1 & 5 / 2 & 1 / 5 & 5 / 2 & 2 / 1 & 5 / 1 \\
$ngram\_range$  & (2, 4) / (2, 5) & (1, 3) / (3, 5) & (1, 3) / (1, 2) & (1, 2) / (2, 4) & (2, 4) / (1, 1) & (2, 5) / (1, 5) & (3, 5) / (1, 4) & (1, 4) / (2, 4) \\

         \bottomrule[1.5pt]
    \end{tabular}
    }
    \caption{Hyperparameters for \textbf{SVM} model in each category across baseline, CW, SR, RW variants. Parameters for experiments using \texttt{title} and \texttt{text} fields are separated by a slash ( / ). Default scikit-learn $tokenizer$ is used when not specified and $analyzer$ is word.} 
    \label{tab:hyperparameters:svm}
\end{table*}

\begin{table*}[h]
\small
    \scalebox{0.80}{
    \begin{tabular}{lcccccccc}
        \multicolumn{9}{c}{\textbf{Hyperparameters for LR}}\\
    \toprule[1.5pt]

& \multicolumn{4}{c}{\textbf{hazard-category}} & \multicolumn{4}{c}{\textbf{product-category}}\\
        \textbf{Parameters}& \textbf{baseline} & \textbf{CW} & \textbf{SR} & \textbf{RW}& \textbf{baseline} & \textbf{CW} & \textbf{SR} & \textbf{RW} \\   & title / text & title / text & title / text & title / text& title / text & title / text & title / text & title / text \\\midrule
$C$ & 5 / 10 & 10 & 10 / 5 & 10 & 10 / 5 & 5 / 10 & 10 / 5 & 10 / 5 \\
$max\_iter$ & 5000 / 1000 & 1000 / 100 & 1000 & 5000 / 1000 & 100 / 5000 & 5000 / 100 & 5000 & 100 / 1000 \\
$max\_features$ & 10000 & 50000 / 10000 & 10000 / 50000 & 50000 / 10000 & 50000 & 50000 & 10000 / 50000 & 50000 \\
$analyzer$ & char / word & word & char & char & char & char & word / char & word \\
$tokenizer$ & - & SpaCy / - & - & - & - & - & - & SpaCy / - \\
$max\_df$ & 0.5 & 0.1 / 0.3 & 0.5 & 0.1 / 0.5 & 0.5 / 0.1 & 0.1 & 0.5 / 0.1 & 0.5 / 0.3 \\
$min\_df$ & 1 / 5 & 2 & 2 / 1 & 2 / 1 & 1 / 2 & 5 & 5 / 2 & 5 / 1 \\
$ngram\_range$ & (3, 5) / (1, 3) & (1, 3) / (1, 1) & (2, 4) / (3, 5) & (3, 5) / (1, 5) & (3, 5) / (1, 4) & (1, 5) / (2, 5) & (1, 2) / (2, 5) & (1, 4) / (1, 1) \\

         \bottomrule[1.5pt]
          & \multicolumn{4}{c}{\textbf{hazard}} & \multicolumn{4}{c}{\textbf{product}}\\
        \textbf{Parameters}& \textbf{baseline} & \textbf{CW} & \textbf{SR} & \textbf{RW}& \textbf{baseline} & \textbf{CW} & \textbf{SR} & \textbf{RW} \\   & title / text & title / text & title / text & title / text& title / text & title / text & title / text & title / text \\\midrule
       $C$ & 10 / 5 & 10 & 10 / 5 & 5 / 10 & 10 & 10 / 5 & 10 / 5 & 5 / 10 \\
       $max\_iter$ & 100 / 1000 & 100 / 5000 & 100 & 100 / 5000 & 1000 / 5000 & 5000 / 100 & 1000 / 5000 & 1000 \\
       $max\_features$ & 10000 / 50000 & 10000 / 5000 & 50000 / 5000 & 50000 / 5000 & 50000 & 50000 / 10000 & 50000 / 5000 & 50000 / 5000 \\
$analyzer$ & char & char & char & char / word & char & word / char & char / word & char \\
$tokenizer$ & - & - & - & - / SpaCy & - & SpaCy / - & - / SpaCy & - \\
$max\_df$ & 0.1 / 0.3 & 0.5 / 0.1 & 0.5 & 0.1 & 0.1 / 0.3 & 0.3 / 0.1 & 0.3 / 0.1 & 0.3 / 0.1 \\

$max\_iter$ & 100 / 1000 & 100 / 5000 & 100 & 100 / 5000 & 1000 / 5000 & 5000 / 100 & 1000 / 5000 & 1000 \\
$min\_df$ & 1 & 5 / 2 & 5 & 1 / 2 & 1 / 5 & 1 / 2 & 1 & 1 / 2 \\
$ngram\_range$ & (2, 4) & (2, 4) / (1, 4) & (1, 4) / (2, 4) & (2, 5) / (1, 1) & (2, 4) / (3, 5) & (1, 1) / (2, 3) & (2, 3) / (1, 1) & (3, 5) / (2, 3) \\

         \bottomrule[1.5pt]
    \end{tabular}
    }
    \caption{Hyperparameters for \textbf{LR} model in each category across baseline, CW, SR, RW variants. Parameters for experiments using \texttt{title} and \texttt{text} fields are separated by a slash ( / ). Default scikit-learn $tokenizer$ is used when not specified and $analyzer$ is word.} 
    \label{tab:hyperparameters:lr}
\end{table*}
\begin{table*}[h]
\small
    \scalebox{0.82}{
    \begin{tabular}{lcccccccc}
        \multicolumn{9}{c}{\textbf{Hyperparameters for DT}}\\
    \toprule[1.5pt]

& \multicolumn{4}{c}{\textbf{hazard-category}} & \multicolumn{4}{c}{\textbf{product-category}}\\
        \textbf{Parameters}& \textbf{baseline} & \textbf{CW} & \textbf{SR} & \textbf{RW}& \textbf{baseline} & \textbf{CW} & \textbf{SR} & \textbf{RW} \\   & title / text & title / text & title / text & title / text& title / text & title / text & title / text & title / text \\\midrule
$max\_depth$ & 100 / 300 & 100 / 200 & 300 / 100 & 200 / 100 & 100 & 200 / 100 & 100 / 300 & 200 / 300 \\
$max\_features$ & 5000 / 50000 & 5000 / 50000 & 50000 & 50000 / 10000 & 50000 / 10000 & 50000 & 10000 & 10000 / 5000 \\
$analyzer$ & word / char & word & word / char & word & char / word & word & char / word & word \\
$tokenizer$ & SpaCy / - & SpaCy & - & - & - & SpaCy / - & - / SpaCy & - \\
$max\_df$ & 0.5 / 0.1 & 0.5 & 0.1 / 0.3 & 0.1 & 0.1 / 0.5 & 0.1 & 0.3 / 0.1 & 0.1 \\
$min\_df$ & 5 / 1 & 5 & 1 / 5 & 1 & 1 & 5 / 2 & 1 / 2 & 5 / 1 \\
$ngram\_range$ & (1, 3) / (2, 5) & (1, 4) / (1, 5) & (1, 3) / (2, 5) & (1, 4) / (1, 1) & (1, 4) / (1, 5) & (1, 4) / (1, 5) & (2, 4) / (1, 2) & (1, 4) / (1, 2) \\
         \bottomrule[1.5pt]
          & \multicolumn{4}{c}{\textbf{hazard}} & \multicolumn{4}{c}{\textbf{product}}\\
        \textbf{Parameters}& \textbf{baseline} & \textbf{CW} & \textbf{SR} & \textbf{RW}& \textbf{baseline} & \textbf{CW} & \textbf{SR} & \textbf{RW} \\   & title / text & title / text & title / text & title / text& title / text & title / text & title / text & title / text \\\midrule
$max\_depth$ & 300 / 100 & 200 / 300 & 200 & 200 & 300 & 100 / 300 & 200 & 200 / 300 \\
$max\_features$ & 50000 / 5000 & 5000 & 5000 / 10000 & 1000 / 50000 & 50000 & 1000 & 1000 & 1000 \\
$analyzer$ & char / word & char / word & word & word & word & char / word & char & word / char \\
$tokenizer$ & - & - / SpaCy & - & - / SpaCy & - / SpaCy & - / SpaCy & - & SpaCy / - \\
$max\_df$ & 0.5 / 0.3 & 0.1 / 0.5 & 0.1 / 0.3 & 0.3 & 0.1 & 0.5 / 0.1 & 0.5 / 0.1 & 0.3 / 0.1 \\
$min\_df$ & 2 / 5 & 2 & 1 / 5 & 2 & 5 / 1 & 2 / 5 & 2 / 1 & 2 / 5 \\
$ngram\_range$ & (3, 5) / (1, 1) & (1, 5) / (1, 2) & (1, 2) / (1, 1) & (1, 5) & (1, 1) / (2, 3) & (2, 3) & (2, 3) / (2, 5) & (1, 4) / (2, 5) \\

         \bottomrule[1.5pt]
    \end{tabular}
    }
    \caption{Hyperparameters for \textbf{DT} model in each category across baseline, CW, SR, RW variants. Parameters for \texttt{title} and \texttt{text} fields are separated by a slash ( / ) unless they are the same. Default scikit-learn $tokenizer$ is used when not specified and $analyzer$ is word.} 
    \label{tab:hyperparameters:dt}
\end{table*}

\begin{table*}[h]
\small
    \scalebox{0.80}{
    \begin{tabular}{lcccccccc}
        \multicolumn{9}{c}{\textbf{Hyperparameters for RF}}\\
    \toprule[1.5pt]

& \multicolumn{4}{c}{\textbf{hazard-category}} & \multicolumn{4}{c}{\textbf{product-category}}\\
        \textbf{Parameters}& \textbf{baseline} & \textbf{CW} & \textbf{SR} & \textbf{RW}& \textbf{baseline} & \textbf{CW} & \textbf{SR} & \textbf{RW} \\   & title / text & title / text & title / text & title / text& title / text & title / text & title / text & title / text \\\midrule
$max\_depth$ & 5000 / 100 & 100 / 1000 & 5000 / 100 & 5000 / 1000 & 1000 / 100 & 5000 / 1000 & 1000 / 100 & 1000 / 100 \\
$n\_estimators$ & 100 / 300 & 100 & 200 & 300 / 200 & 300 & 300 & 200 & 200 / 300 \\
$max\_features$ & 10000 / 50000 & 10000 / 50000 & 10000 / 50000 & 50000 & 50000 / 10000 & 10000 / 50000 & 10000 & 50000 \\
$analyzer$ & char & word / char & char & char & word & word / char & word & word \\
$tokenizer$ & - & - & - & - & SpaCy & SpaCy / - & - & SpaCy \\
$max\_df$ & 0.3 & 0.1 / 0.3 & 0.1 & 0.1 / 0.3 & 0.1 / 0.3 & 0.1 & 0.3 / 0.1 & 0.1 \\
$min\_df$ & 2 & 1 & 5 / 1 & 5 / 2 & 1 / 2 & 5 / 2 & 2 / 1 & 5 / 2 \\
$ngram\_range$ & (3, 5) / (2, 5) & (1, 5) & (1, 4) / (1, 5) & (3, 5) / (1, 5) & (1, 2) / (1, 1) & (1, 2) / (1, 5) & (1, 5) / (1, 1) & (1, 2) / (1, 3) \\
         \bottomrule[1.5pt]
          & \multicolumn{4}{c}{\textbf{hazard}} & \multicolumn{4}{c}{\textbf{product}}\\
        \textbf{Parameters}& \textbf{baseline} & \textbf{CW} & \textbf{SR} & \textbf{RW}& \textbf{baseline} & \textbf{CW} & \textbf{SR} & \textbf{RW} \\   & title / text & title / text & title / text & title / text& title / text & title / text & title / text & title / text \\\midrule
$max\_depth$ & 5000 / 1000 & 1000 & 1000 / 100 & 1000 / 5000 & 1000 & 1000 & 1000 & 5000 / 1000 \\
$n\_estimators$ & 300 / 200 & 300 / 100 & 200 / 300 & 200 & 300 & 200 & 200 / 100 & 300 / 200 \\
$max\_features$ & 50000 & 10000 / 50000 & 5000 / 10000 & 5000 / 50000 & 50000 / 10000 & 50000 / 5000 & 10000 & 5000 / 50000 \\
$analyzer$ & word / char & char & char & char & char / word & char & word & word \\
$tokenizer$ & SpaCy / - & - & - & - & - / SpaCy & - & SpaCy / - & SpaCy \\

$max\_df$ & 0.1 & 0.5 / 0.1 & 0.5 & 0.5 / 0.3 & 0.3 & 0.3 / 0.1 & 0.1 & 0.3 / 0.1 \\

$min\_df$ & 2 / 1 & 2 / 1 & 5 / 1 & 2 & 2 / 5 & 1 / 5 & 1 / 5 & 2 / 1 \\
$ngram\_range$ & (1, 2) / (2, 5) & (1, 4) / (1, 5) & (2, 4) / (1, 5) & (1, 5) / (3, 5) & (3, 5) / (1, 4) & (2, 5) / (1, 3) & (1, 1) / (1, 2) & (1, 3) / (1, 1) \\

         \bottomrule[1.5pt]
    \end{tabular}
    }
    \caption{Hyperparameters for \textbf{RF} model in each category across baseline, CW, SR, RW variants. Parameters for experiments using \texttt{title} and \texttt{text} fields are separated by a slash ( / ). Default scikit-learn $tokenizer$ is used when not specified and $analyzer$ is word.} 
    \label{tab:hyperparameters:rf}
\end{table*}

\begin{table*}[h]
\small
    \scalebox{0.80}{
    \begin{tabular}{lcccccccc}
        \multicolumn{9}{c}{\textbf{Hyperparameters for KNN}}\\
    \toprule[1.5pt]

& \multicolumn{4}{c}{\textbf{hazard-category}} & \multicolumn{4}{c}{\textbf{product-category}}\\
        \textbf{Parameters}& \textbf{baseline} & \textbf{CW} & \textbf{SR} & \textbf{RW}& \textbf{baseline} & \textbf{CW} & \textbf{SR} & \textbf{RW} \\   & title / text & title / text & title / text & title / text& title / text & title / text & title / text & title / text \\\midrule
$n\_neighbors$ & 3 / 7 & 7 / 3 & 11 / 3 & 5 & 5 & 11 / 5 & 3 / 5 & 5 \\
$weights$ & distance & distance & distance & distance & uniform / distance & distance & distance & distance \\
$analyzer$ & char / word & char / word & char / word & char & word / char & char & char & char \\
$tokenizer$ & - & - & - & - & SpaCy / - & - & - & - \\
$max\_df$ & 0.3 / 0.1 & 0.5 / 0.3 & 0.5 / 0.3 & 0.5 / 0.1 & 0.3 / 0.1 & 0.3 / 0.1 & 0.1 & 0.1 \\
$min\_df$ & 2 / 5 & 2 & 1 & 5 & 1 & 1 / 5 & 5 / 2 & 5 / 1 \\
$ngram\_range$ & (1, 3) / (1, 4) & (1, 3) & (1, 4) & (2, 3) / (2, 4) & (1, 1) / (3, 5) & (1, 5) / (2, 4) & (1, 4) & (1, 5) / (3, 5) \\

         \bottomrule[1.5pt]
          & \multicolumn{4}{c}{\textbf{hazard}} & \multicolumn{4}{c}{\textbf{product}}\\
        \textbf{Parameters}& \textbf{baseline} & \textbf{CW} & \textbf{SR} & \textbf{RW}& \textbf{baseline} & \textbf{CW} & \textbf{SR} & \textbf{RW} \\   & title / text & title / text & title / text & title / text& title / text & title / text & title / text & title / text \\\midrule
$n\_neighbors$ & 7 & 11 / 7 & 9 / 7 & 7 & 3 & 3 & 7 / 5 & 3 \\
$weights$ & distance & distance & distance & distance & distance & uniform / distance & distance & uniform \\
$analyzer$ & word / char & char & char & char & char & char & word / char & char \\
$tokenizer$ & - & - & - & - & - & - & - & - \\
$max\_df$ & 0.1 & 0.3 & 0.3 / 0.1 & 0.3 & 0.3 / 0.5 & 0.5 / 0.1 & 0.1 & 0.1 / 0.3 \\
$min\_df$ & 2 / 5 & 1 & 2 / 5 & 5 / 1 & 5 & 1 / 2 & 5 / 2 & 2 / 5 \\
$ngram\_range$ & (1, 3) / (3, 5) & (2, 5) / (2, 4) & (2, 3) / (2, 4) & (1, 5) & (2, 3) / (1, 5) & (2, 5) & (1, 4) / (2, 5) & (2, 5) \\
         \bottomrule[1.5pt]
    \end{tabular}
    }
    \caption{Hyperparameters for \textbf{KNN} model in each category across baseline, CW, SR, RW variants. Parameters for experiments using \texttt{title} and \texttt{text} fields are separated by a slash ( / ). Default scikit-learn $tokenizer$ is used when not specified and $analyzer$ is word.} 
    \label{tab:hyperparameters:knn}
\end{table*}

\begin{table*}[h]
\small
    \scalebox{0.86}{
    \begin{tabular}{lcccccccc}
        \multicolumn{9}{c}{\textbf{Hyperparameters for NB}}\\
    \toprule[1.5pt]

& \multicolumn{4}{c}{\textbf{hazard-category}} & \multicolumn{4}{c}{\textbf{product-category}}\\
        \textbf{Parameters}& \textbf{baseline} & \textbf{CW} & \textbf{SR} & \textbf{RW}& \textbf{baseline} & \textbf{CW} & \textbf{SR} & \textbf{RW} \\   & title / text & title / text & title / text & title / text& title / text & title / text & title / text & title / text \\\midrule
$alpha$ & 0.01 & 0.01 & 0.01 & 0.01 & 0.01 & 0.01 & 0.1 / 0.01 & 0.1 / 0.01 \\
$analyzer$ & word / char & char / word & char / word & word & char & char / word & word / char & word \\
$tokenizer$ & - & - & - & - / SpaCy & - & - & - & SpaCy \\
$max\_df$ & 0.1 / 0.5 & 0.5 / 0.1 & 0.3 / 0.5 & 0.1 & 0.1 & 0.1 / 0.3 & 0.1 & 0.3 / 0.1 \\
$min\_df$ & 2 / 5 & 2 & 1 / 2 & 1 / 2 & 2 & 2 & 1 / 5 & 2 / 1 \\
$ngram\_range$ & (1, 3) / (2, 5) & (3, 5) / (2, 4) & (1, 4) / (2, 3) & (2, 5) / (1, 3) & (3, 5) / (1, 4) & (2, 5) / (1, 1) & (1, 2) / (3, 5) & (1, 1) \\

         \bottomrule[1.5pt]
          & \multicolumn{4}{c}{\textbf{hazard}} & \multicolumn{4}{c}{\textbf{product}}\\
        \textbf{Parameters}& \textbf{baseline} & \textbf{CW} & \textbf{SR} & \textbf{RW}& \textbf{baseline} & \textbf{CW} & \textbf{SR} & \textbf{RW} \\   & title / text & title / text & title / text & title / text& title / text & title / text & title / text & title / text \\\midrule
$alpha$ & 0.01 & 0.01 & 0.1 / 0.01 & 0.01 & 0.01 / 0.1 & 0.01 & 0.1 & 0.1 / 0.01 \\
$analyzer$ & char & word & word / char & char & char / word & char & char & word / char \\
$tokenizer$ & - & - & - & - & - / SpaCy & - & - & - \\
$max\_df$ & 0.1 / 0.3 & 0.5 / 0.1 & 0.5 / 0.1 & 0.3 / 0.5 & 0.1 / 0.5 & 0.3 / 0.1 & 0.1 & 0.1 \\
$min\_df$ & 1 / 5 & 2 & 2 / 5 & 1 & 5 & 1 & 5 / 2 & 1 \\
$ngram\_range$ & (2, 4) / (3, 5) & (1, 1) / (2, 4) & (1, 2) / (1, 5) & (2, 5) / (3, 5) & (2, 5) / (1, 1) & (2, 5) / (2, 4) & (1, 3) / (2, 5) & (1, 1) / (1, 3) \\

         \bottomrule[1.5pt]
    \end{tabular}
    }
    \caption{Hyperparameters for \textbf{NB} model in each category across baseline, CW, SR, RW variants. Parameters for experiments using \texttt{title} and \texttt{text} fields are separated by a slash ( / ). Default scikit-learn $tokenizer$ is used when not specified and $analyzer$ is word.} 
    \label{tab:hyperparameters:nb}
\end{table*}

\begin{table*}[h]
\small
    \scalebox{0.85}{
    \begin{tabular}{lcccccccc}
        \multicolumn{9}{c}{\textbf{Hyperparameters for BERT}}\\
    \toprule[1.5pt]
& \multicolumn{4}{c}{\textbf{hazard-category}} & \multicolumn{4}{c}{\textbf{product-category}}\\
        \textbf{Parameters}& \textbf{baseline} & \textbf{CW} & \textbf{SR} & \textbf{RW}& \textbf{baseline} & \textbf{CW} & \textbf{SR} & \textbf{RW} \\   & title / text & title / text & title / text & title / text& title / text & title / text & title / text & title / text \\\midrule

         $batch\_size$ & 8 / 32 & 32 / 16 & 16 / 8 & 16 / 32 & 16 / 8 & 32 / 8 & 8 / 32 & 32 \\
$epochs$ & 5 / 10 & 5 / 3 & 10 & 3 & 5 / 10 & 5 & 5 / 3 & 3 / 5 \\
$lr\_scheduler$ & cosRestarts / cos & lin & lin & lin / cos & cosRestarts & cosRestarts / lin & cosRestarts / lin & cosRestarts \\
         \bottomrule[1.5pt]
          & \multicolumn{4}{c}{\textbf{hazard}} & \multicolumn{4}{c}{\textbf{product}}\\
        \textbf{Parameters}& \textbf{baseline} & \textbf{CW} & \textbf{SR} & \textbf{RW}& \textbf{baseline} & \textbf{CW} & \textbf{SR} & \textbf{RW} \\   & title / text & title / text & title / text & title / text& title / text & title / text & title / text & title / text \\\midrule
       $batch\_size$ & 16 / 8 & 16 / 8 & 16 & 8 & 16 / 8 & 32 & 32 / 16 & 32 \\
$epochs$ & 10 & 10 / 3 & 3 / 5 & 3 / 5 & 10 & 3 / 10 & 5 & 3 / 5 \\
$lr\_scheduler$ & lin / cos & lin / cos & lin & cos / lin & lin & cosRestarts / cos & cos / cosRestarts & lin / cosRestarts \\

         \bottomrule[1.5pt]
    \end{tabular}
    }
    \caption{Hyperparameters for \textbf{BERT} model in each category across baseline, CW, SR, RW variants. Parameters for experiments using \texttt{title} and \texttt{text} fields are separated by a slash ( / ). Learning rate schedulers: cos (cosine annealing), cosRestarts (cosine annealing with restarts), and lin (linear).} 
    \label{tab:hyperparameters:bert}
\end{table*}

\begin{table*}[h]
\small
    \scalebox{0.78}{
    \begin{tabular}{lcccccccc}
        \multicolumn{9}{c}{\textbf{Hyperparameters for RoBERTa}}\\
    \toprule[1.5pt]

& \multicolumn{4}{c}{\textbf{hazard-category}} & \multicolumn{4}{c}{\textbf{product-category}}\\
        \textbf{Parameters}& \textbf{baseline} & \textbf{CW} & \textbf{SR} & \textbf{RW}& \textbf{baseline} & \textbf{CW} & \textbf{SR} & \textbf{RW} \\   & title / text & title / text & title / text & title / text& title / text & title / text & title / text & title / text \\\midrule
       $batch\_size$ & 8 / 32 & 16 / 32 & 8 / 16 & 32 / 16 & 32 / 16 & 8 / 32 & 32 & 16 \\
$epochs$ & 3 / 10 & 10 & 10 / 5 & 10 / 5 & 5 / 10 & 3 / 5 & 3 / 10 & 10 / 3 \\
$lr\_scheduler$ & lin / cos & lin / cosRestarts & cosRestarts / lin & cosRestarts & cosRestarts & lin / cosRestarts & lin & cos \\

         \bottomrule[1.5pt]
          & \multicolumn{4}{c}{\textbf{hazard}} & \multicolumn{4}{c}{\textbf{product}}\\
        \textbf{Parameters}& \textbf{baseline} & \textbf{CW} & \textbf{SR} & \textbf{RW}& \textbf{baseline} & \textbf{CW} & \textbf{SR} & \textbf{RW} \\   & title / text & title / text & title / text & title / text& title / text & title / text & title / text & title / text \\\midrule
       $batch\_size$ & 16 & 32 / 16 & 32 & 16 / 32 & 16 / 32 & 16 & 32 & 32 / 16 \\
$epochs$ & 10 & 10 & 3 & 3 / 5 & 5 / 10 & 5 & 5 & 10 \\
$lr\_scheduler$ & lin / cosRestarts & lin & cos / cosRestarts & cos / lin & cosRestarts / cos & cosRestarts / cos & cosRestarts / cos & lin \\

         \bottomrule[1.5pt]
    \end{tabular}
    }
    \caption{Hyperparameters for \textbf{RoBERTa} model in each category across baseline, CW, SR, RW variants. Parameters for experiments using \texttt{title} and \texttt{text} fields are separated by a slash ( / ). Learning rate schedulers: cos (cosine annealing), cosRestarts (cosine annealing with restarts), and lin (linear).} 
    \label{tab:hyperparameters:roberta}
\end{table*}

\begin{table*}[h]
\small
    \scalebox{0.85}{
    \begin{tabular}{lcccccccc}
        \multicolumn{9}{c}{\textbf{Hyperparameters for DistilBERT}}\\
    \toprule[1.5pt]

& \multicolumn{4}{c}{\textbf{hazard-category}} & \multicolumn{4}{c}{\textbf{product-category}}\\
        \textbf{Parameters}& \textbf{baseline} & \textbf{CW} & \textbf{SR} & \textbf{RW}& \textbf{baseline} & \textbf{CW} & \textbf{SR} & \textbf{RW} \\   & title / text & title / text & title / text & title / text& title / text & title / text & title / text & title / text \\\midrule
       $batch\_size$ & 8 & 32 / 16 & 16 & 16 / 8 & 16 / 8 & 8 & 16 / 32 & 16 / 32 \\
$epochs$ & 10 / 5 & 10 / 5 & 10 / 5 & 10 / 3 & 3 / 10 & 5 & 5 & 5 / 3 \\
$lr\_scheduler$ & cos & cosRestarts & cos / lin & lin / cosRestarts & lin / cosRestarts & lin & cos / cosRestarts & cos \\

         \bottomrule[1.5pt]
          & \multicolumn{4}{c}{\textbf{hazard}} & \multicolumn{4}{c}{\textbf{product}}\\
        \textbf{Parameters}& \textbf{baseline} & \textbf{CW} & \textbf{SR} & \textbf{RW}& \textbf{baseline} & \textbf{CW} & \textbf{SR} & \textbf{RW} \\   & title / text & title / text & title / text & title / text& title / text & title / text & title / text & title / text \\\midrule
       $batch\_size$ & 8 & 8 / 16 & 32 & 32 & 8 / 32 & 32 / 16 & 16 & 16 / 32 \\
$epochs$ & 10 & 3 / 5 & 5 & 3 / 10 & 10 & 10 & 3 / 10 & 5 \\
$lr\_scheduler$ & lin & cos / lin & cosRestarts / cos & lin / cosRestarts & cos / cosRestarts & cos & cosRestarts / lin & lin \\

         \bottomrule[1.5pt]
    \end{tabular}
    }
    \caption{Hyperparameters for \textbf{DistilBERT} model in each category across baseline, CW, SR, RW variants. Parameters for experiments using \texttt{title} and \texttt{text} fields are separated by a slash ( / ). Learning rate schedulers: cos (cosine annealing), cosRestarts (cosine annealing with restarts), and lin (linear).} 
    \label{tab:hyperparameters:distilbert}
\end{table*}

\begin{table*}[h]
\small
    \scalebox{0.76}{
    \begin{tabular}{lcccccccc}
        \multicolumn{9}{c}{\textbf{Hyperparameters for ModernBERT}}\\
    \toprule[1.5pt]

& \multicolumn{4}{c}{\textbf{hazard-category}} & \multicolumn{4}{c}{\textbf{product-category}}\\
        \textbf{Parameters}& \textbf{baseline} & \textbf{CW} & \textbf{SR} & \textbf{RW}& \textbf{baseline} & \textbf{CW} & \textbf{SR} & \textbf{RW} \\   & title / text & title / text & title / text & title / text& title / text & title / text & title / text & title / text \\\midrule
       $batch\_size$ & 16 / 8 & 8 / 32 & 16 & 8 / 32 & 32 / 8 & 8 / 32 & 16 / 8 & 16 / 32 \\
$epochs$ & 3 / 5 & 5 & 5 / 3 & 10 / 5 & 5 / 10 & 5 / 10 & 5 & 10 \\
$lr\_scheduler$ & cos & cosRestarts / lin & cos / lin & cosRestarts / cos & lin / cos & cos / cosRestarts & cosRestarts / lin & cos / cosRestarts \\

         \bottomrule[1.5pt]
          & \multicolumn{4}{c}{\textbf{hazard}} & \multicolumn{4}{c}{\textbf{product}}\\
        \textbf{Parameters}& \textbf{baseline} & \textbf{CW} & \textbf{SR} & \textbf{RW}& \textbf{baseline} & \textbf{CW} & \textbf{SR} & \textbf{RW} \\   & title / text & title / text & title / text & title / text& title / text & title / text & title / text & title / text \\\midrule
       $batch\_size$ & 32 / 8 & 16 / 8 & 8 & 32 / 8 & 8 & 8 & 8 & 8 \\
$epochs$ & 10 & 5 / 10 & 5 & 5 & 10 & 5 & 10 / 5 & 3 \\
$lr\_scheduler$ & cosRestarts / lin & cos / cosRestarts & lin / cosRestarts & cos & cos & cosRestarts & lin / cos & cos \\

         \bottomrule[1.5pt]
    \end{tabular}
    }
    \caption{Hyperparameters for \textbf{ModernBERT} model in each category across baseline, CW, SR, RW variants. Parameters for experiments using \texttt{title} and \texttt{text} fields are separated by a slash ( / ). Learning rate schedulers: cos (cosine annealing), cosRestarts (cosine annealing with restarts), and lin (linear). } 
    \label{tab:hyperparameters:modernbert}
\end{table*}

\end{document}